arXiv:1506.04838v1 [cs.LG] 16 Jun 2015

# Spectral Sparsification and Regret Minimization Beyond Matrix Multiplicative Updates*

Zeyuan Allen-Zhu
zeyuan@csail.mit.edu
MIT CSAIL

Zhenyu Liao
zhenyul@bu.edu
Boston University

Lorenzo Orecchia
orecchia@bu.edu
Boston University

March 09, 2015

**Abstract**

In this paper, we provide a novel construction of the linear-sized spectral sparsifiers of Batson, Spielman and Srivastava [BSS14]. While previous constructions required $\Omega(n^4)$ running time [BSS14, Zou12], our sparsification routine can be implemented in almost-quadratic running time $O(n^{2+\varepsilon})$.

The fundamental conceptual novelty of our work is the leveraging of a strong connection between sparsification and a regret minimization problem over density matrices. This connection was known to provide an interpretation of the randomized sparsifiers of Spielman and Srivastava [SS11] via the application of matrix multiplicative weight updates (MWU) [CHS11, Vis14]. In this paper, we explain how matrix MWU naturally arises as an instance of the Follow-the-Regularized-Leader framework and generalize this approach to yield a larger class of updates. This new class allows us to accelerate the construction of linear-sized spectral sparsifiers, and give novel insights on the motivation behind Batson, Spielman and Srivastava [BSS14].

# 1 Introduction

A powerful tool to handle large-scaled graphs is to compress them by reducing their sizes, while preserving properties of interest such as the size of cuts [BK96, BK02] or the routability of certain flows [CLLM10]. This *sparsification* procedures also play an important role as fundamental primitives behind many fast graph algorithms [KLOS14, PS14]. In this paper, we consider the strong notion of *spectral sparsifier* put forward by Spielman and Teng [ST04, ST11]: $G'$ is $(1+\varepsilon)$-spectral approximate to $G$ if $G'$ is a subgraph of $G$ with possibly reweighted edges, and for every $x \in \mathbb{R}^n$,

$$x^T L_G x \le x^T L_{G'} x \le (1+\varepsilon) x^T L_G x \quad \text{or equivalently} \quad L_G \preceq L_{G'} \preceq (1+\varepsilon) L_G \ ,$$

where $L_G$ and $L_{G'}$ are respectively the graph Laplacian matrices of $G$ and $G'$.

The algorithm of Spielman and Srivastava [SS11] constructs $(1+\varepsilon)$-spectral sparsifiers with $O(n \log n/\varepsilon^2)$ edges in nearly linear time by randomly sampling edges proportionally to their effective resistance. In a seminal paper, Batson, Spielman and Srivastava [BSS14] give $(1+\varepsilon)$-spectral sparsifiers with $O(n/\varepsilon^2)$ edges, but their construction and subsequent algorithm by [Zou12] require $O(mn^3/\varepsilon^2)$ and $O(mn^2/\varepsilon^2 + n^4/\varepsilon^4)$ time respectively. We shall refer to their analysis and algorithm the BSS for short. The main contribution of this paper is to give an improved construction of linear-sized spectral sparsifiers that runs in almost-quadratic time.

**Theorem 1.** *For any even integer $q \ge 2$ and any $\varepsilon \in (0, \frac{1}{4\sqrt{q}})$, there is an algorithm that, for any weighted undirected graph $G$ with $n$ vertices and $m$ edges, with probability at least $1 - n^{-\Omega(1)}$, constructs a $(1+\varepsilon)$-spectral sparsifier $G'$ that has at most $O(\sqrt{q}n/\varepsilon^2)$ edges in time $\widetilde{O}(mn^{1+1/q}/\varepsilon^5)$.*

Since $q$ can be chosen as a large constant and the graph can be preprocessed to reduce the number of edges to $m = O(n \log n)$, the above running time is almost quadratic in terms of $n$.

Graph sparsification is a special case of sparsifying sums of rank-1 PSD matrices (see [BSS14] and Appendix B). Our algorithm for Theorem 1 also applies to this more general problem with an almost cubic running time, which is stil an improvement over the previous quartic running time.

**Theorem 2.** *For any even integer $q \ge 2$ and any $\varepsilon \in (0, \frac{1}{4\sqrt{q}})$, there is an algorithm that, for any decomposition $I = \sum_{i=1}^m v_i v_i^T \in \mathbb{R}^{n \times n}$ of rank-1 matrices, with probability at least $1 - n^{-\Omega(1)}$, constructs scalars $s_i \ge 0$ with $|\{i : s_i > 0\}| \le O(\sqrt{q}n/\varepsilon^2)$ that satisfies $I \preceq \sum_{i=1}^m s_i v_i v_i^T \preceq (1+\varepsilon) I$ in time $\widetilde{O}(n^{3+1/q}/\varepsilon^5 + mn/\varepsilon^4)$.*

The fundamental conceptual novelty of our work is the establishment of a deep connection between graph or matrix sparsifications and a regret minimization problem over PSD matrices (see Section 1.1). This relation was known [CHS11, Vis14] for the randomized sparsifiers of Spielman and Srivastava [SS11], for which the underlying matrix concentration bound can be easily recovered as an application of the matrix version of Multiplicative Weight Updates (MWU) [AK07, Ore11], a standard online learning algorithm. However, it was not clear how this interpretation could be extended to BSS, despite a clear analogy was also noted by de Carli Silva, Harvey and Sato (see [CHS11, Section 8]). Both the MWU and the BSS rely on potential function arguments, where the potential is essentially a robust version to capture of the maximum and minimum graph eigenvalues. In this paper, we provide the missing piece of this interpretation: we consider a generalization of MWU to a larger class of updates, and show that the BSS can be recovered as an instance of this class. Beyond our faster implementation of sparsification, we believe that this interpretation is of independent interest and may be useful in other areas in which the argument of BSS has found application [Nao12].

We focus on updates coming from the *follow-the-regularized-leader* (FTRL) framework. The choice of regularizer in this framework fully determines the update strategy and the corresponding potential function. See for example the recent survey by Hazan [Haz12]. The standard MWU argument can be recovered as an instance of FTRL, where the regularizer is chosen to be the entropy function. In contrast, we choose a different class of regularizers consisting of all $\ell_{1-1/q}$ semi-norms for $q \geq 2$, and provide corresponding regret bounds in Section 3. In Section 4 and Section 5, we show that the choice $q = 2$ recovers an algorithm which is somewhat similar to BSS, and produces linear-sized spectral sparsifiers. This algorithm can be implemented to run in a $O(mn^{3/2})$ time. Finally, in Section 6, we consider regularizers corresponding to large, constant $q > 2$, which yield very different algorithms from BSS with almost quadratic running time.

### 1.1 Regret Minimization

In this subsection, we discuss our contribution on the problem of regret minimization in online linear optimization [Haz12]. Our technical results apply to the more general case of online PSD linear optimization over the set of density matrices, but our key contributions are described more concisely in the scalar case.

Let $\Delta_n = \{x \in \mathbb{R}^n : x \geq 0 \wedge \mathbb{1}^T x = 1\}$ be the unit simplex in $\mathbb{R}^n$, and we call a vector in $\Delta_n$ an *action*. A player is going to play $T$ actions $x_0, \ldots, x_{T-1} \in \Delta_n$ in a row; only after playing $x_k$, the player observes a feedback vector $f_k \in \mathbb{R}^n$, which may depend on $x_k$, and suffers the linear loss $\langle f_k, x_k \rangle$. The regret minimization problem asks us to device a strategy for the player that minimizes the *regret*, i.e., difference between the total loss suffered by the player and the loss suffered by the *a* posteriori best fixed action $u \in \Delta_n$:

$$\text{minimize} \quad \max_{u \in \Delta_n} R(u), \qquad \text{where } R(u) \stackrel{\text{def}}{=} \textstyle\sum_{k=0}^{T-1} \langle f_k, x_k - u \rangle \ .$$

A well-known strategy for this problem is to update $x_k$ in a multiplicative fashion: for each coordinate $i \in [n]$, define $x_{k+1,i}$ to be proportional to $x_{k,i} \cdot \exp^{-\alpha \cdot f_{k,i}}$ for some parameter $\alpha > 0$. This strategy is known as the *multiplicative weight update*. Its classical analysis [AHK12] implies

$$\forall u \in \Delta_n, \quad R(u) = \sum_{k=0}^{T-1} \langle f_k, x_k - u \rangle \leq \frac{\alpha}{2} \sum_{k=0}^{T-1} \|f_k\|_\infty^2 + \frac{\log n}{\alpha} \ . \tag{1.1}$$

The first term on the righthand side contributes a regret of $\|f_k\|_\infty^2$ that is paid at every iteration, and we call it the *width term.* The second term is a fixed start-up cost corresponding to 'how long it takes the update to explore the whole $\Delta_n$', and we call it the *diameter term.* If for all iterations $k$, $\|f_k\|_\infty$ is upper bounded by $\rho$, known as the *width* of the problem, the trade-off between the width and diameter terms can be be optimized by the choice of $\alpha > 0$ to show that the total regret is at most $O(\rho\sqrt{T \log n})$.

**Optimization Interpretation.** We take an optimization perspective to describe MWU and its generalizations by characterizing our strategies as instances of the *follow-the-regularized-leader* and *mirror descent* frameworks. Let $w(\cdot)$ be a strongly convex function over the simplex, known as the *regularizer.* The follow-the-regularized-leader strategy with parameter $\alpha > 0$ can be described as a trade-off between minimizing the loss incurred so far and the value of the regularizer.

$$\texttt{FTRL:} \qquad x_{k+1} = \arg\min_{z \in \Delta_n} \left\{ w(z) + \alpha \textstyle\sum_{j=0}^{k} \langle f_j, z \rangle \right\} \ . \tag{1.2}$$

Similarly, the mirror-descent strategy optimizes a trade-off

$$\texttt{MirrorDescent:} \qquad \text{start with } x_0 = \left(\tfrac{1}{n}, \ldots, \tfrac{1}{n}\right); \quad x_{k+1} \leftarrow \arg\min_{z \in \Delta_n} \{V_{x_k}(z) + \alpha \langle f_k, z \rangle\} \ , \tag{1.3}$$

where $V_x(y) \stackrel{\text{def}}{=} w(y) - w(x) - \langle \nabla w(x), y - x \rangle$ is the induced Bregman divergence. Under mild assumptions (which are satisfied in this paper, see Appendix A), it is easy to check that `MirrorDescent` is equivalent to `FTRL`. We will therefore interchangeably use `MirrorDescent` and `FTRL` in the rest of the paper, because `FTRL` gives the cleaner description for the updates, while `MirrorDescent` provides a simpler analysis. The MWU strategy is an instance of the two equivalent strategies above, with the choice of regularizer $w(x) \stackrel{\text{def}}{=} \sum_i x_i \log x_i - x_i$, i.e. the (negative) entropy function.

**Previous Work.** The MWU is a simple but extremely powerful algorithmic tool that has been repeatedly discovered in theory of computation, machine learning, optimization, and game theory (see for instance the survey [AHK12] and the book [CL06]). Since MWU has found numerous important applications in semidefinite programming [AK07, AHK05], constraint satisfaction problem [Ste10], maximum flow [CKM$^+$11], sparsest cut [She09], balanced separator [OSV12], small set expansion [BFK$^+$11], traveling salesman problem [AGM$^+$10], zero-sum games [DDK11], and fractional packing problems [GK07]. The analysis of follow-the-regularized-leader can be found in the surveys [Haz12, Sha07], while that of the mirror descent appears in the the book [BN13].

**Beyond MWU.** Historically, MWU has been extended at least from three orthogonal directions. In this paper, we pursue all these three directions simultaneously (see our summary in Table 1.)

1. **From vector to matrix.** Instead of studying actions $x$ in the forms of $n$-dimensional probability distributions, one can study density matrices $X$ in $\Delta_{n\times n}$, the set of PSD matrices whose trace equals to one. This is a generalization from a set of "experts" corresponding to $\{\mathbf{e}_1, \dots, \mathbf{e}_n\}$ to all combinations of the form $\sum_{i=1}^n t_i \mathbf{e}_i$ where $t$ is on the $n$-dimensional unit sphere $\mathbb{S}^{n-1}$. Accordingly, each loss vector $f_k$ can be generalized to a symmetric matrix $F_k \in \mathbb{R}^{n\times n}$, so the loss of any density matrix $X$ becomes $F_k \bullet X = \mathrm{Tr}(F_k X)$. (If $X = vv^T$ is of rank one, then $F_k \bullet X = v^T F_k v$.) Among many applications, the matrix version of MWU has been used in designing algorithms for solving semidefinite programs [AK07] and finding balanced separators [OSV12], and in the proof of $\mathsf{QIP} = \mathsf{PSPACE}$ [JJUW11].
2. **Local norm convergence.** The width term $\|f_k\|_\infty^2$ in the regret upper bound (1.1) can be replaced with $\langle |f_k|, x_k\rangle \cdot \|f_k\|_\infty$. (Here, we have used $|f_k|$ to denote coordinate-wise absolute value of $f_k$.) This technique is known as the local-norm technique because $\langle |f_k|, x_k\rangle$ is a local way to measure the length of $f_k$ with respect to $x_k$. Since $\langle |f_k|, x_k\rangle \cdot \|f_k\|_\infty$ is never larger than $\|f_k\|_\infty^2$, as well as $x_k \in \Delta_n$, this new upper bound can only be smaller than the original. Indeed, this tighter bound has proved useful in the multi-arm bandit problem [AHR12], and in the solution of positive linear programs [AO15]. It also underpins the negative-width technique of [AHK12].
3. **Change of regularizer.** If one replaces the entropy regularizer with the $\ell_{1-1/q}$-regularizer $w(x) = -\frac{q}{q-1}\sum_{i=1}^n x_i^{1-1/q}$ for any $q \geq 2$, the corresponding update rule changes
$$\text{from} \quad \boxed{x_{k+1,i} = \exp^{-\sum_{j=0}^k \alpha f_{j,i} + c}} \quad \text{to} \quad \boxed{x_{k+1,i} = \left(\sum_{j=0}^k \alpha f_{j,i} + c\right)^{-q}} \ ,$$
where in both cases $c$ is the unique constant that ensures $x_{k+1} \in \Delta_n$. The FTRL framework is very powerful as the choice of regularizer $w(x)$ completely determines both the form and the analysis of the update strategy. Ultimately, different regularizers achieve different trade-offs between the width and diameter terms in Equation (1.1). For instance, the $\ell_{1/2}$-regularizer yields the following regret bound
$$\forall u \in \Delta_n, \quad R(u) \leq O(\alpha) \cdot \sum_{k=0}^{T-1} \langle |f_k|, x_k\rangle \cdot \max_{i\in[n]} |f_{k,i}\sqrt{x_{k,i}}| + \frac{2\sqrt{n}}{\alpha} \ .$$
The diameter term is now $2\sqrt{n}$, much worse than $\log n$ in the entropy case in (1.1). However, since (the local norm version of) the width term goes from $\langle |f_k|, x_k\rangle \cdot \|f_k\|_\infty$ to $\langle |f_k|, x_k\rangle \cdot \max_{i\in[n]} |f_{k,i}\sqrt{x_{k,i}}|$, the width term may become smaller.. This is exactly the case in the sparsification case, where the feedback vectors, corresponding to the edges added to the sparsifier, may be weighted up by a factor as large as $n$, so that we may have $\|f_k\|_\infty \geq n$. In this scenario, the use of a more stongly-convex regularizer, such as $\ell_{1/2}$, allows us to measure the width in a more convenient local norm and yields the BSS linear-sized sparsifier(see Figure 1 on page 12 for a visual comparison of different regularizers). We point out that the $\ell_{1-1/q}$-regularizers have also been used, albeit solely in the scalar case, by the machine learning community to obtain asymptotically optimal strategies for the multi-arm bandit problem [ABL11, BC12].

| Paper | Allow Matrix? | Allow Local Norm? | Allow Non-Entropy Regularizer? |
|---|---|---|---|
| [PST95, FS95] [AHK05, AHK12] | no | no | no |
| [AHR12, AO15] | no | yes | no |
| [ABL11, BC12] | no | yes | yes |
| [AK07, OSV12] | yes | no | no |
| [HKS12] | yes | yes | no |
| **[this paper]** | yes | yes | yes |

Table 1: Comparisons among prior results on the regret minimization problem.

### 1.2 Extensions

**High Rank Sparsification.** Our same algorithm of Theorem 1 and 2 also applies to sparsifying sums of PSD matrices, rather than just rank-1 PSD matrices. This recovers the same result of de Carli Silva, Harvey, and Sato [CHS11]. Such an extension has been shown important for problems such as finding hypergraph sparsifiers, finding sparse SDP solutions, and finding sparsifiers on subgraphs. However, as in the rank-1 case, the detailed running time of our algorithm has to be examined separately for each specific sparsification problem.

As an example, given a weighted undirected graph $G$ that is decomposed into edge-disjoint subgraphs, the goal of *linear-sized subgraph sparsification* is to construct a $(1+O(\varepsilon))$-spectral sparsifier $G'$ to $G$, so that $G'$ consists only of the reweighted versions of at most $n/\varepsilon^2$ given subgraphs. Our same algorithm for Theorem 1 runs in time $\widetilde{O}(mn^{1+1/q}/\varepsilon^5)$ for this problem.

**Weak Unweighted Graph Sparsification.** Given $\kappa \in [1, m/n]$, consider the problem of finding a $\kappa$-spectral sparsifier of $G$ containing $O(m/\kappa)$ distinct edges from $E$, *without* reweighting. This problem is very recently studied by Anderson, Gu and Melgaard [AGM14], our regret minimization framework allows us to design a simple and almost-quadratic-time algorithm for this problem, improving from the quartic time complexity of [AGM14].

## 2 Preliminaries

Throughout this paper, for a cleaner representation that depends on the context, we interchangeably use $X \bullet Y = \langle X, Y\rangle = \mathrm{Tr}(XY)$ to denote the inner product between two symmetric matrices. If $X$ is symmetric, we use $e^X$ to denote its matrix exponential and $\log X$ to denote its matrix logarithm, when $X$ is PSD. If $X$ is symmetric with eigendecomposition $X = \sum_{i=1}^n \lambda_i v_i v_i^T$ we denote by $|X| \stackrel{\text{def}}{=} \sum_{i=1}^n |\lambda_i| v_i v_i^T$. For any symmetric $X$, we use $\|X\|_{\mathsf{spe}}$ to denote the spectral norm of $X$, and $\lambda_{\mathsf{max}}(X), \lambda_{\mathsf{min}}(X)$ to denote its largest and smallest eigenvalues. We define $\Delta_{n\times n} \stackrel{\text{def}}{=} \{X \in \mathbb{R}^{n\times n} : X \succeq 0, \mathrm{Tr}X = 1\}$ to be the set of positive semidefinite (PSD) matrices with trace 1. This should be seen as the matrix generalization of the $n$-dimensional simplex $\Delta_n \stackrel{\text{def}}{=} \{x \in \mathbb{R}^n : x \geq 0, \mathbb{1}^T x = 1\}$.

**Regularizers and Bregman Divergence.** We are interested in two types of regularizers over $\Delta_{n\times n}$, namely, $w(X) \stackrel{\text{def}}{=} X \bullet (\log X - I)$, known as the entropy regularizer, and $w(X) \stackrel{\text{def}}{=} -\frac{q}{q-1}\text{Tr}X^{1-1/q}$ for some $q > 1$, which we call the $\ell_{1-1/q}$-regularizer. The corresponding Bregman divergences $V_X(Y) \stackrel{\text{def}}{=} w(Y) - w(X) - \langle \nabla w(X), Y - X\rangle$ are the following.

$$\text{entropy case:} \quad V_X(Y) = Y \bullet (\log Y - \log X) - I \bullet (Y - X) \ ,$$

$$\ell_{1-1/q} \text{ case:} \quad V_X(Y) = X^{-1/q} \bullet Y + \frac{1}{q-1}\text{Tr}X^{1-1/q} - \frac{q}{q-1}\text{Tr}Y^{1-1/q} \ .$$

Note that both regularizers above and their Bregman divergences are convex over the cone of PSD matrices.[1] We now state some classical properties of Bregman divergence. Their proofs are included in Appendix D for completeness.

**Lemma 2.1.** *The Bregman divergence of a convex differentiable function $w(\cdot)$ has the properties:*

- *Non-negativity: $V_X(Y) \geq 0$ for all $X, Y \geq 0$.*
- *The "three-point equality": $\langle \nabla w(X) - \nabla w(Y), X - U\rangle = V_X(U) - V_Y(U) + V_Y(X)$.*
- *Given $\widetilde{X} \succeq 0$ and $X = \arg\min_{Z\in\Delta_{n\times n}} V_{\widetilde{X}}(Z)$ as the* Bregman projection*, we have the "generalized Pythagorean theorem" for all $U \in \Delta_{n\times n}$: $V_{\widetilde{X}}(U) \geq V_X(U) + V_{\widetilde{X}}(X) \geq V_X(U)$.*

# 3 Regret Minimization in Full Information

In this section, we consider the following setting of the regret minimization problem, known as the full information setting. At each iteration $k = 0, \ldots, T-1$, the player chooses an action $X_k \in \Delta_{n\times n}$, receives a symmetric loss matrix $F_k \in \mathbb{R}^{n\times n}$ and suffers a loss $\langle F_k, X_k\rangle$. At this point, the player is allowed to observe the full matrix $F_k$ without any restriction.

Again, the goal of the player is to minimize the regret with respect to any fixed matrix $U \in \Delta_{n\times n}$:

$$R(U) \stackrel{\text{def}}{=} \textstyle\sum_{k=0}^{T-1}\langle F_k, X_k - U\rangle \ .$$

The best choice of $U$ in hindsight can be taken as the rank-1 projection over a minimum eigenvector of $\sum_{k=0}^{T-1} F_k$. As a result, the total loss for the best choice of $U$ is $\lambda_{\mathsf{min}}\big(\sum_{k=0}^{T-1} F_k\big)$.

**Entropy Regularizer.** If $w(\cdot)$ is the entropy regularizer, then (1.2) can be explicitly written as

$$\texttt{MirrorDescent}_{\mathsf{exp}}: \qquad X_k = \exp^{cI - \alpha\sum_{j=0}^{k-1} F_j} \ , \tag{3.1}$$

where $c \in \mathbb{R}$ is the unique constant that ensures $\text{Tr}X_k = 1$. This is also known as the *matrix multiplicative weight update* method, and the following theorem gives its regret bound.[2]

**Theorem 3.1.** *In $\texttt{MirrorDescent}_{\mathsf{exp}}$, if the parameter $\alpha > 0$ satisfies $\alpha F_k \succeq -I$ for all iterations $k = 0, 1, \ldots, T-1$, then, for every $U \in \Delta_{n\times n}$,*

$$R(U) \stackrel{\text{def}}{=} \sum_{k=0}^{T-1}\langle F_k, X_k - U\rangle \leq \alpha\sum_{k=0}^{T-1}\big(X_k \bullet |F_k|\big)\cdot\|F_k\|_{\mathsf{spe}} + \frac{V_{X_0}(U)}{\alpha} \ .$$

*We note that $V_{X_0}(U) \leq \log n$.*

Our proof of Theorem 3.1 uses a technique known as the *tweaked version* of mirror descent (see [Zin03, Rak09]). We define an intermediate point $\widetilde{X}_{k+1} = \arg\min_{Z\succeq 0}\big\{V_{X_k}(Z) + \alpha\langle F_k, Z\rangle\big\}$ as the minimizer over $Z \succeq 0$, rather than $Z \in \Delta_{n\times n}$ as in (1.3). Accordingly, the actual point $X_{k+1}$ equals to $\arg\min_{Z\in\Delta_{n\times n}}\{V_{\widetilde{X}_{k+1}}(Z)\}$, the *Bregman projection* of $\widetilde{X}_{k+1}$ back to the hyperplane $\mathrm{Tr}Z = 1$. This two-step interpretation of mirror descent gives a very clean proof to our regret bound, and we defer this full proof to Appendix E.

[1] While this is easy to check by taking the second derivative for the entropy regularizer, it is less obvious for the $\ell_{1-1/q}$ regularizer. The latter follows easily from Lieb's concavity theorem [Lie73, Bha97].

[2] The scalar version of this theorem was proved for instance in [AR09, Sha11, AO15]. A slightly different matrix version of this theorem was proved in [HKS12] (in particular, the authors of [HKS12] have required $I \succeq \alpha F_k \succeq -I$ while in fact it suffices to only require $\alpha F_k \succeq -I$.

**$\ell_{1-1/q}$ regularizer.** If $w(\cdot)$ is the $\ell_{1-1/q}$ regularizer, then (1.2) can be explicitly written as

$$\texttt{MirrorDescent}_{\ell_{1-1/q}}: \qquad X_k = \Big(cI + \alpha \textstyle\sum_{j=0}^{k-1} F_j\Big)^{-q} , \tag{3.2}$$

where $c \in \mathbb{R}$ is the unique constant that ensures $cI + \alpha\sum_{j=0}^{k-1} F_j \succ 0$ and $\mathrm{Tr}X_k = 1$.

If we focus on the special case of $q = 2$ and each $F_k$ having rank 1, the following theorem gives the regret bound for $\texttt{MirrorDescent}_{\ell_{1/2}}$.

**Theorem 3.2.** *In* $\texttt{MirrorDescent}_{\ell_{1/2}}$*, if the parameter* $\alpha > 0$*, and the loss matrix* $F_k$ *is rank one and satisfies* $X_k^{1/2} \bullet \alpha F_k > -1$ *for all* $k$*, then, for every* $U \in \Delta_{n\times n}$*,*

$$R(U) \stackrel{\text{def}}{=} \sum_{k=0}^{T-1}\langle F_k, X_k - U\rangle \le \alpha \cdot \sum_{k=0}^{T-1} \frac{(X_k \bullet F_k)(X_k^{1/2} \bullet F_k)}{1 + X_k^{1/2} \bullet \alpha F_k} + \frac{V_{X_0}(U)}{\alpha} \enspace .$$

*If we instead have* $X_k^{1/2} \bullet \alpha F_k \ge -\frac{1}{2}$*, the above bound can be simplified as*

$$R(U) \stackrel{\text{def}}{=} \sum_{k=0}^{T-1}\langle F_k, X_k - U\rangle \le 2\alpha \cdot \sum_{k=0}^{T-1} (X_k \bullet F_k)(X_k^{1/2} \bullet F_k) + \frac{V_{X_0}(U)}{\alpha} \enspace .$$

*We note that* $V_{X_0}(U) \le 2\sqrt{n}$*.*

We recommend the interested readers to see the proof of Theorem 3.2 in Appendix E, as it provides a straightforward generalization of Theorem 3.1 using regularizers other than entropy.

Theorem 3.2 is only a special case of the following more general regret bound, which holds for arbitrary $q \ge 2$, and for $F_k$ having arbitrary rank. At a first reading, one can skip Theorem 3.3 because its sole purpose in this paper is to improve the running time of graph sparsification from $\widetilde{O}(mn^{3/2})$ to $\widetilde{O}(mn^{1+1/q})$, as well as allowing one to sparsify sums of high rank PSDs.

**Theorem 3.3.** *In* $\texttt{MirrorDescent}_{\ell_{1-1/q}}$ *with* $q \ge 2$ *and* $\alpha > 0$*, if the loss matrix* $F_k$ *is either positive or negative semidefinite and satisfies* $\alpha X_k^{1/2q} F_k X_k^{1/2q} \succeq -\frac{1}{2q} I$ *for all* $k$*, then for every* $U \in \Delta_{n\times n}$*,*

$$R(U) \stackrel{\text{def}}{=} \sum_{k=0}^{T-1}\langle F_k, X_k - U\rangle \le O(q\alpha) \sum_{k=0}^{T-1} (X_k \bullet |F_k|) \cdot \|X_k^{1/2q} F_k X_k^{1/2q}\|_{\mathsf{spe}} + \frac{V_{X_0}(U)}{\alpha} \enspace .$$

*We note that* $V_{X_0}(U) \le \frac{q}{q-1} n^{1/q}$*.*

(The proof of Theorem 3.3 is deferred to Appendix E.)

The key idea to prove Theorem 3.3 is to replace the use of the Sherman-Morrison formula in the proof of Theorem 3.2 with the Woodbury formula so as to allow $F_k$ to be of high rank. It also uses the Lieb-Thirring trace inequality to handle arbitrary $q \ge 2$.)

# 4 Warm-Up: Upper-Sided Linear-Sized Sparsification

In this section and the next, we present our construction of linear-sized sparisifier in the general matrix setting. Its specialization to graph sparsification appears in Appendix B, while its efficient implementation is discussed in Section 6. To showcase how the regret bounds of Section 3 can be useful in the construction of sparsifiers, we start by describing a warm-up example in which we are only interested in obtaining a single side of the sparsification guarantee.

Suppose we are given a decomposition of the identity matrix $I = \sum_{e=1}^m w_e \widehat{L}_e$, where each $\widehat{L}_e$ satisfies

$$0 \preceq \widehat{L}_e \preceq I \text{ and is of rank 1 and trace 1, i.e. } \widehat{L}_e = vv^t \text{ for some } v \in \mathbb{R}^n \text{ with } \|v\|_2 = 1.$$

The weights $w_e > 0$ may be unknown, though the trace guarantee ensures that $\sum_e w_e = n$. In this section, we are interested in finding some $s \in \Delta_m$ satisfying $\sum_{e=1}^m (ns_e) \cdot \widehat{L}_e \preceq (1+\varepsilon)I$, while the sparsity of $s$ —that is, $|\{e \in [m] : s_e > 0\}|$— is at most $O(n/\varepsilon^2)$. We call this the *upper-sided linear-sized spectral sparsification* because it only gives an upper bound on the eigenvalues of $\sum_{e=1}^m (ns_e) \cdot \widehat{L}_e$ and no lower bound.

Consider the following algorithm that invokes the regret minimization framework in Section 3 to solve this upper-sided sparsification. We choose

$$\text{the } \ell_{1/2} \text{ regularizer and } \alpha = \varepsilon/4\sqrt{n} \text{ for } \texttt{MirrorDescent}_{\ell_{1/2}}.$$

At iteration $k$, set the feedback matrix as $F_k = -n\widehat{L}_{e_k}$, where $e_k$ minimizes $\widehat{L}_e \bullet X_k$ over $e \in [m]$. [3]

Before applying Theorem 3.2, let us first verify that the prerequisite $X_k^{1/2} \bullet \alpha F_k \geq -\frac{1}{2}$ holds. Because $\sum_{e\in[m]} \frac{w_e}{n} \widehat{L}_e \bullet X_k = \frac{1}{n} I \bullet X_k = \frac{1}{n}$, by an averaging argument, we must have $\widehat{L}_{e_k} \bullet X_k \leq \frac{1}{n}$. This further implies $-\alpha n \widehat{L}_{e_k} \bullet X_k^{1/2} \geq -\alpha\sqrt{n} > -\frac{1}{2}$ due to the claim below.

**Claim 4.1.** *For every $X \in \Delta_{n\times n}$, we have $\widehat{L}_e \bullet X^{1/2} \leq (\widehat{L}_e \bullet X)^{1/2}$ for every $e \in [m]$.*

*Proof.* Without loss of generality, one can assume $X$ to be diagonal. Next, since $\widehat{L}_e = v_e v_e^T$ is of rank one, the desired inequality follows from Jensen's inequality $v_e^T X^{1/2} v_e \leq \sqrt{v_e^T X v_e}$ and the fact that $\|v_e\|_2^2 = \mathrm{Tr}\widehat{L}_e \leq 1$. □

Now, applying Theorem 3.2, we obtain that for every $U \in \Delta_{n\times n}$,

$$\sum_{k=0}^{T-1} \langle -n\widehat{L}_{e_k}, X_k - U\rangle \leq 2\alpha \cdot \sum_{k=0}^{T-1} (X_k \bullet n\widehat{L}_{e_k})(X_k^{1/2} \bullet n\widehat{L}_{e_k}) + \frac{2\sqrt{n}}{\alpha} \quad .$$

After rearranging, and using $\widehat{L}_{e_k} \bullet X_k \leq \frac{1}{n}$ and $n\widehat{L}_{e_k} \bullet X_k^{1/2} \leq \sqrt{n}$ we deduced earlier,

$$\begin{aligned}\Big\langle \frac{n}{T} \sum_{k=0}^{T-1} \widehat{L}_{e_k}, U \Big\rangle &\leq \frac{2\alpha}{T} \cdot \sum_{k=0}^{T-1} (X_k \bullet n\widehat{L}_{e_k})(X_k^{1/2} \bullet n\widehat{L}_{e_k}) + \frac{1}{T}\sum_k \langle n\widehat{L}_{e_k}, X_k\rangle + \frac{2\sqrt{n}}{\alpha T} \\ &\leq \frac{2\alpha}{T} \cdot T \cdot 1 \cdot \sqrt{n} + 1 + \frac{2\sqrt{n}}{\alpha T} = \frac{\varepsilon}{2} + 1 + \frac{8n}{\varepsilon T} \quad .\end{aligned}$$

Finally, choosing $T = 16n/\varepsilon^2$ and $U$ to be the rank-1 projection over a maximum eigenvector, we conclude that $\lambda_{\max}(\frac{n}{T}\sum_{k=0}^{T-1} \widehat{L}_{e_k}) \leq 1 + \varepsilon$.

This completes the description of our upper-sided linear-sized sparsification algorithm. The full sparsification algorithm, in the next section, will essentially consists of playing out this analysis on the lower and upper side at the same time.

We emphasize here that if one chooses the entropy regularizer by using $\texttt{MirrorDescent}_{\texttt{exp}}$, and chooses $e_k = e$ with probability proportional to $w_e$, a similar analysis from the one above recovers the sparsification result of Spielman and Srivastava [SS11].

---

[3] This choice naturally follows from a saddle-point interpretation of the problem, because it is the subgradient of the function $f(X) \stackrel{\text{def}}{=} \min_{s\in\Delta_m} \sum_{e=1}^m (ns_e \widehat{L}_e) \bullet X$ at $X = X_k$. We have skipped the explanation of this choice due to the space limitation.

## 5 Linear-Sized Sparsification

As before, suppose we are given a decomposition of the identity matrix $I = \sum_{e=1}^m w_e \widehat{L}_e$, where each $\widehat{L}_e$ satisfies $0 \preceq \widehat{L}_e \preceq I$ and is of rank 1 and trace 1. The weights $w_e > 0$ may be unknown and satsify $\sum_e w_e = n$. In this section, we are interested in finding scalars $s_e \geq 0$ satisfying

$$I \preceq \textstyle\sum_{e=1}^m s_e \cdot \widehat{L}_e \preceq (1 + 8\varepsilon + O(\varepsilon^2))I \ , \tag{5.1}$$

while the sparsity of $s$ —that is, $|\{e \in [m] : s_e > 0\}|$— is at most $O(n/\varepsilon^2)$.

Instead of maintaining one sequence $X_k$ like in Section 4, we maintain two sequences $X_k, Y_k \in \Delta_{n\times n}$. At each iteration $k \in 0, 1, \ldots, T-1$, find an arbitrary $e_k \in [m]$ such that

$$\widehat{L}_{e_k} \bullet X_k \leq \widehat{L}_{e_k} \bullet Y_k \ .$$

This is always possible by an averaging argument with weights $w_e$. Next, we choose the $\ell_{1/2}$ regularizer and some parameter $\alpha < 1/2$ (in fact, we will choose $\alpha = \varepsilon$ later), and updates

$$X_{k+1} = \operatorname*{arg\,min}_{Z \in \Delta_{n\times n}} \Big\{ V_{X_k}(Z) + \Big\langle \frac{-\alpha \widehat{L}_{e_k}}{(X_k \bullet \widehat{L}_{e_k})^{1/2}}, Z \Big\rangle \Big\} \quad \text{and}$$

$$Y_{k+1} = \operatorname*{arg\,min}_{Z \in \Delta_{n\times n}} \Big\{ V_{Y_k}(Z) + \Big\langle \frac{\alpha \widehat{L}_{e_k}}{(Y_k \bullet \widehat{L}_{e_k})^{1/2}}, Z \Big\rangle \Big\} \ . \tag{5.2}$$

In other words, we have picked feedback matrices $F_k = \frac{-\widehat{L}_{e_k}}{(X_k \bullet \widehat{L}_{e_k})^{1/2}}$ for the $X_k$ sequence and $F_k = \frac{\widehat{L}_{e_k}}{(Y_k \bullet \widehat{L}_{e_k})^{1/2}}$ for the $Y_k$ sequence in our $\texttt{MirrorDescent}_{\ell_{1/2}}$.[4]

Notice that $X_k^{1/2} \bullet \frac{-\alpha \widehat{L}_{e_k}}{(X_k \bullet \widehat{L}_{e_k})^{1/2}} \geq -\frac{1}{2}$ due to Claim 4.1, so we always have $X_k^{1/2} \bullet \alpha F_k \geq -\frac{1}{2}$ which satisfies the prerequisite of Theorem 3.2. Applying Theorem 3.2 on the $X_k$ sequence, we obtain that for every $U_X \in \Delta_{n\times n}$,

$$\begin{aligned}\sum_{k=0}^{T-1} \Big\langle \frac{-\widehat{L}_{e_k}}{(X_k \bullet \widehat{L}_{e_k})^{1/2}}, X_k - U_X \Big\rangle &\leq 2\alpha \cdot \sum_{k=0}^{T-1} \Big(X_k \bullet \frac{\widehat{L}_{e_k}}{(X_k \bullet \widehat{L}_{e_k})^{1/2}}\Big)\Big(X_k^{1/2} \bullet \frac{\widehat{L}_{e_k}}{(X_k \bullet \widehat{L}_{e_k})^{1/2}}\Big) + \frac{V_{X_0}(U_X)}{\alpha} \\ &= 2\alpha \cdot \sum_{k=0}^{T-1} X_k^{1/2} \bullet \widehat{L}_{e_k} + \frac{V_{X_0}(U_X)}{\alpha} \leq 2\alpha \cdot \sum_{k=0}^{T-1} (X_k \bullet \widehat{L}_{e_k})^{1/2} + \frac{V_{X_0}(U_X)}{\alpha} \ .\end{aligned}$$

Above, the last inequality uses Claim 4.1. If we denote by $M_X \stackrel{\text{def}}{=} \sum_{k=0}^{T-1} \frac{\widehat{L}_{e_k}}{(\widehat{L}_{e_k} \bullet X_k)^{1/2}}$ and rearrange the inequality above, we get

$$M_X \bullet U_X \leq \frac{V_{X_0}(U_X)}{\alpha} + (1 + 2\alpha) \sum_{k=0}^{T-1} (\widehat{L}_{e_k} \bullet X_k)^{1/2} \ . \tag{5.3}$$

Similarly, applying Theorem 3.2 on the $Y_k$ sequence, and define $M_Y \stackrel{\text{def}}{=} \sum_{k=0}^{T-1} \frac{\widehat{L}_{e_k}}{(\widehat{L}_{e_k} \bullet Y_k)^{1/2}}$, we obtain that for every $U_Y \in \Delta_{n\times n}$,

$$M_Y \bullet U_Y \geq -\frac{V_{Y_0}(U_Y)}{\alpha} + (1 - 2\alpha) \sum_{k=0}^{T-1} (\widehat{L}_{e_k} \bullet Y_k)^{1/2} \ . \tag{5.4}$$

In the rest of the proof, we will use (5.3) and (5.4) to deduce

$$\lambda_{\mathsf{max}}(M_Y) - \lambda_{\mathsf{min}}(M_Y) \leq 8\varepsilon(1 + O(\varepsilon))\lambda_{\mathsf{min}}(M_Y) \ . \tag{5.5}$$

[4] In fact, the denominator $(X_k \bullet \widehat{L}_{e_k})^{1/2}$ is defined so as to make sure that $F_k$ is the 'maximally aggressive' loss matrix we can have for $\texttt{MirrorDescent}_{\ell_{1/2}}$.

Finally, since $M_Y = \sum_{k=0}^{T-1} \frac{\widehat{L}_{e_k}}{(\widehat{L}_{e_k} \bullet Y_k)^{1/2}}$ is a matrix that is a summation of at most $T = n/\varepsilon^2$ rank-1 matrices, dividing it by $\lambda_{\mathsf{min}}(M_Y)$ gives the desired sparsification for (5.1).

We prove (5.5) in two steps.

**Lowerbounding $\lambda_{\mathsf{min}}(M_Y)$.** Recall that we have $\mathrm{Tr}(M_X) = \sum_{k=0}^{T-1} \frac{1}{(\widehat{L}_e \bullet X_k)^{1/2}}$ because we have assumed each $\widehat{L}_e$ to be of trace 1. Denoting by $a_k = (\widehat{L}_e \bullet X_k)^{1/2}$, we have that $\mathrm{Tr}(M_X) = \sum_{k=0}^{T-1} \frac{1}{a_k}$. We apply (5.3) here with $U_X = \frac{1}{n} I = X_0$, and obtain

$$\frac{1}{n} \sum_{k=0}^{T-1} \frac{1}{a_k} = \frac{1}{n} \mathrm{Tr}(M_X) \leq (1+2\alpha) \sum_{k=0}^{T-1} (\widehat{L}_e \bullet X_k)^{1/2} \leq (1+2\alpha) \sum_{k=0}^{T-1} a_k \enspace .$$

Applying Cauchy-Schwarz, we have

$$(\sum_{k=0}^{T-1} a_k)^2 \geq \frac{1}{n(1+2\alpha)} (\sum_{k=0}^{T-1} a_k)(\sum_{k=0}^{T-1} \frac{1}{a_k}) \geq \frac{T^2}{n(1+2\alpha)} \enspace . \tag{5.6}$$

If we choose $T = \frac{n}{\varepsilon^2}$, we immediately have[5]

$$\textstyle\sum_{k=0}^{T-1} (\widehat{L}_e \bullet Y_k)^{1/2} \geq \sum_{k=0}^{T-1} a_k \geq \frac{\sqrt{n}}{\varepsilon^2}(1 - O(\alpha)) \enspace .$$

Substituting the above lower bound into (5.4), and choosing $U_Y \in \Delta_{n \times n}$ to be the rank-1 projection matrix over the smallest eigenvector of $M_Y$, and choosing $\alpha = \varepsilon$, we have

$$\lambda_{\mathsf{min}}(M_Y) \geq -\frac{2\sqrt{n}}{\alpha} + (1-2\alpha) \sum_{k=0}^{T-1} (\widehat{L}_e \bullet Y_k)^{1/2} \geq (1 - O(\varepsilon)) \frac{\sqrt{n}}{\varepsilon^2} \tag{5.7}$$

**Upperbounding $\lambda_{\mathsf{max}}(M_Y) - \lambda_{\mathsf{min}}(M_Y)$.** This time, we use our choice of $\widehat{L}_{e_k} \bullet X_k \leq \widehat{L}_{e_k} \bullet Y_k$ to combine (5.3) and (5.4) and derive that

$$\frac{1}{1+2\alpha} M_Y \bullet U_X \leq \frac{1}{1+2\alpha} M_X \bullet U_X \leq \frac{1}{1-2\alpha} M_Y \bullet U_Y + \frac{2\sqrt{n}}{\alpha} (\frac{1}{1+2\alpha} + \frac{1}{1-2\alpha}) \enspace .$$

Choosing $U_X$ to be the rank-1 matrix projection matrix over the largest eigenvector of $M_Y$, $U_Y$ to be that over the smallest eigenvector of $M_Y$, and recalling that $\alpha = \varepsilon$, we have

$$\lambda_{\mathsf{max}}(M_Y) \leq \frac{1+2\varepsilon}{1-2\varepsilon} \lambda_{\mathsf{min}}(M_Y) + \frac{4\sqrt{n}}{\varepsilon}(1 + O(\varepsilon)) \enspace .$$

After rearranging and substituting in the lower bound (5.7), we finish the proof of (5.5)

$$\lambda_{\mathsf{max}}(M_Y) - \lambda_{\mathsf{min}}(M_Y) \leq \frac{4\varepsilon}{1-2\varepsilon} \lambda_{\mathsf{min}}(M_Y) + \frac{4\sqrt{n}}{\varepsilon}(1 + O(\varepsilon)) \leq 8\varepsilon(1 + O(\varepsilon)) \lambda_{\mathsf{min}}(M_Y) \enspace . \qquad \square$$

# 6 Efficient Implementation for Graph Sparsification

The update rules described in (5.2) imply that $X_k$ and $Y_k$ are of the form (see Section 3)

$$X_k = \Big(c^X \cdot I - \textstyle\sum_{j=0}^{k-1} s_j^X \widehat{L}_{e_j}\Big)^{-2} \quad \text{and} \quad Y_k = \Big(\textstyle\sum_{j=0}^{k-1} s_j^Y \widehat{L}_{e_j} - c^Y \cdot I\Big)^{-2} \enspace . \tag{6.1}$$

Here, $c^X$ is the unique (positive) constant that satisfies $c^X I - \sum_{j=0}^{k-1} s_j^X \widehat{L}_{e_j} \succ 0$ and $\mathrm{Tr} X_k = 1$, while $c^Y$ is the unique (possibly negative) constant that satisfies $\sum_{j=0}^{k-1} s_j^Y \widehat{L}_{e_j} - c^Y I \succ 0$ and $\mathrm{Tr} Y_k = 1$.

[5] In fact, it suffices to stop our algorithm at the earliest iteration $T$ so that inequality (5.6) is satisfied. Our analysis here only represents the most pessimistic scenario; in practice, this early termination implies we can choose less than $n/\varepsilon^2$ matrices for certain inputs. This is in contrast to [BSS14], as their algorithm uses $n/\varepsilon^2$ rank-1 matrices for all inputs.

The coefficients $s_j^X$ and $s_j^Y$ are always positive. (It is worth noting that $c^X$ is initially $\sqrt{n}$ at $X_0$ and keeps increasing, while $c^Y$ is initially $-\sqrt{n}$ and keeps increasing as well.)

Recall that $\texttt{MirrorDescent}_{\ell_{1/2}}$ requires one to compute $c^X$ and $c^Y$ for each iteartion, and this can be done via binary search. One way to perform binary search is to first compute $\lambda_{\max} = \lambda_{\max}(\sum_{j=0}^{k-1} s_j^X \widehat{L}_{e_j})$. Then, one can binary search $c^X$ in the range $[\lambda_{\max}+1, \lambda_{\max}+\sqrt{n}]$ to find the correct one satisfying $\mathrm{Tr}\big(c^X \cdot I - \sum_{j=0}^{k-1} s_j^X \widehat{L}_{e_j}\big)^{-2} = 1$. Similarly, one can binary search $c^Y$ in the range of $[\lambda_{\min} - \sqrt{n}, \lambda_{\min} - 1]$ where $\lambda_{\min} = \lambda_{\min}(\sum_{j=0}^{k-1} s_j^Y \widehat{L}_{e_j})$.[6]

If one performs the binary search to an accuracy that is small enough, this gives an algorithm whose running time is $\widetilde{O}(n^3 m/\varepsilon^2)$, dominated by the computation of $X_k \bullet \widehat{L}_e = \big(c^X \cdot I - \sum_{j=0}^{k-1} s_j^X \widehat{L}_{e_j}\big)^{-2} \bullet \widehat{L}_e$ for each $k \in [T]$ and $e \in [m]$.

**Running Time Improvement.** For the graph sparsification problem described in Theorem 1, we sketch the key ideas needed to improve the running time to $\widetilde{O}\big(mn^{1+1/q}/\varepsilon^5\big)$ for any even integer $q \geq 2$. The details can be found in Appendix F and G. In particular, we first describe how to achieve a running time of $\widetilde{O}\big(mn^{1+1/2}/\varepsilon^5\big)$.

Recall that in Section 5, we have constructed $M_X$ and $M_Y$ and proved that $\lambda_{\min}(M_X)$ and $\lambda_{\min}(M_Y)$ are both at least $\Omega(\sqrt{n}/\varepsilon^2)$. In fact, it is not hard to ensure that $\lambda_{\max}(M_X)$ and $\lambda_{\max}(M_Y)$ are at most $O(\sqrt{n}/\varepsilon^2)$ as well.[7] Since $\sum_{j=0}^{k-1} s_j^X \widehat{L}_{e_j} \preceq \alpha M_X$, we conclude that the eigenvalues of $\sum_{j=0}^{k-1} s_j^X \widehat{L}_{e_j}$ are all upper bounded by $\alpha \cdot O(\sqrt{n}/\varepsilon^2) = O(\sqrt{n}/\varepsilon)$. Therefore, throughout the algorithm, the encountered choices of $c^X$ are always upper bounded by $O(\sqrt{n}/\varepsilon)$.

For this reason, we only need to compute matrix inversions of the form $(cI - A)^{-1}$, with the guarantee that $c = O(\sqrt{n}/\varepsilon)$. Since we always have $cI - A \succeq I$ —as otherwise $\mathrm{Tr}(cI-A)^{-2}$ is strictly larger than 1— we can approximate this matrix inverse by

$$(cI - A)^{-1} = c^{-1}\Big(I - \frac{A}{c}\Big)^{-1} \approx c^{-1}\Big(I + \frac{A}{c} + \frac{A^2}{c^2} + \cdots \frac{A^d}{c^d}\Big) \ , \tag{6.2}$$

and it suffices to choose the maximum degree $d = O(\sqrt{n}/\varepsilon)$. This is formally proved in Lemma G.6. In other words, when computing $X_k$, it suffices to replace the matrix inversion with some matrix polynomial of degree $d = O(\sqrt{n}/\varepsilon)$. Similar idea also holds for the $Y_k$ sequence.

So far, we managed avoiding the computationally expensive matrix inversion. Next, we want to further accelerate the procedure of computing $(cI - A)^{-2} \bullet \widehat{L}_e$ for all edges $e \in [m]$ simultaneously. Recall that $\widehat{L}_e = v_e v_e^T$ is of rank 1, and one can rewrite

$$(cI - A)^{-2} \bullet \widehat{L}_e = v_e^T (cI - A)^{-2} v_e = \|(cI - A)^{-1} v_e\|_2^2 \ .$$

For this reason, as in [SS11], one can apply the Johnson-Lindenstrauss dimension reduction [JL84]: there exists random matrix $Q$ with $\widetilde{O}(1/\varepsilon^2)$ rows, satisfying that $\|(cI - A)^{-1} v_e\|_2^2 \approx \|Q(cI - A)^{-1} v_e\|_2^2$ for for all $v_e$.

Using this dimension reduction, one can precompute $T = Q(cI - A)^{-1}$ in time $\widetilde{O}(m/\varepsilon^2) \times \widetilde{O}(\sqrt{n}/\varepsilon) = \widetilde{O}(m\sqrt{n}/\varepsilon^3)$, with the help from the approximate matrix inversion (6.2), and the nearly-linear time Laplacian system solvers [ST04]. After the precomputation, each $(cI - A)^{-2} \bullet \widehat{L}_e \approx \|T v_e\|_2^2$ can be computed in $\widetilde{O}(1/\varepsilon^2)$ time, totaling $\widetilde{O}(m/\varepsilon^2)$ per iteration, which is negligible.

In sum, taking into account that we have $T = n/\varepsilon^2$ iterations, the total running time is $\widetilde{O}(mn^{1+1/2}/\varepsilon^5)$. To turn this $\widetilde{O}(mn^{1+1/2}/\varepsilon^5)$ into $\widetilde{O}(mn^{1+1/q}/\varepsilon^5)$ for any constant $q$, we need to replace the use of the $\ell_{1/2}$ regularizer with the $\ell_{1-1/q}$ regularizer. This requires one to use Theorem 3.3 in replacement of Theorem 3.2.

[6] $\lambda_{\max}$ and $\lambda_{\min}$ can be computed via power methods, and it suffices to compute them up to an additive error of, say, 0.1. In Appendix G, we propose an alternative approach to compute $c^X$ and $c^Y$, avoiding the use of power methods.

[7] This may require one to stop the algorithm earlier than $T = n/\varepsilon^2$ iterations, which is even better!

We wish to emphasize here that our analysis in Section 5 needs to be *strengthened* in order to tolerate all the errors incurred from the approximate computations (most notably from Laplacian linear solvers, from Johnson-Lindenstrauss, and from (6.2)). This is only rountinary thanks to the optimization motivation behind our argument, and we have done this carefully in Appendix F.

## Acknowledgement

We thank Richard Peng, Nikhil Srivastava, and Nisheeth Vishnoi for helpful conversations. This material is based upon work partly supported by the National Science Foundation under Grant CCF-1319460 and by a Simons Graduate Student Award under grant no. 284059.

# Appendix

**Appendix roadmap.**

- In Figure 1, we plot the entropy and the $\ell_{1/2}$ regularizers of the 3-dimensional scalar case for a visual comparison.
- In Appendix A, we verify the equivalence between `FTRL` and `MirrorDescent` for our choices of the regularizers.
- In Appendix B, we provide notations for graphs, and state the reduction from the sparsifying graphs to sparsifying sums of rank-1 matrices.
- In Appendix C, we provide our unweighted sparsification result.
- In Appendix D and E we provide missing proofs for Section 2 and 3 respectively.
- In Appendix F, we generalize our sparsification algorithm of Section 5 to allow arbitrary $q \geq 2$, high rank matrices, and approximate computations.
- In Appendix G, we provide the details of how to implement linear-sized graph sparsifications in almost-quadratic time, thus finishing the running time claim of Theorem 1.
- In Appendix H, we sketch how to generalize our running time improvement to other problems, including sparsifying sums of rank-1 PSD matrices (i.e., Theorem F.5), as well as subgraph sparsifications.

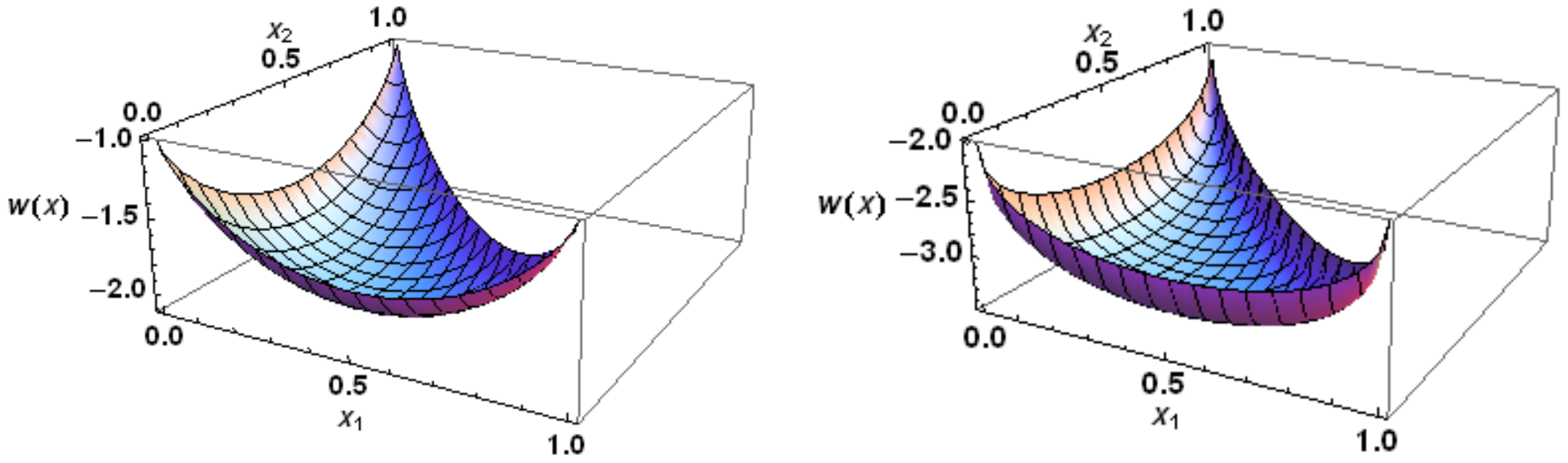

(a) The entropy regularizer

(b) The $\ell_{1/2}$ regularizer

Figure 1: Two regularizers in $n = 3$. The first two axes represent $x_1, x_2$ so $x_3 = 1 - x_1 - x_2$. The third axes represent $w(x)$.

## A Partial Equivalence Between FTRL and Mirror Descent

In this section, we show the equivalence between mirror descent and follow-the-regularized-leader for our choices of the regularizers. In fact, this equivalence holds more generally for all regularizers $w(\cdot)$ that are convex function of Legendre type with domain $Q$ (see for instance [BMDG05, Roc96]).

Letting $A_i \in \mathbb{R}^n$ be any symmetric matrix for each iteration $i$, the follow-the-regularized-leader method can be described as

$$\forall k = 0, 1, \ldots, T-1, \quad X_k = \operatorname*{arg\,min}_{Z \in \Delta_{n \times n}} \Big\{ w(Z) + \sum_{i=0}^{k-1} \langle A_i, Z \rangle \Big\} \ . \tag{A.1}$$

The mirror descent method (with starting point $\widetilde{X}_0 = \frac{1}{n} I$) can be described as

$$\forall k = 0, 1, \ldots, T-1, \quad \widetilde{X}_k = \operatorname*{arg\,min}_{Z \in \Delta_{n \times n}} \Big\{ V_{\widetilde{X}_{k-1}}(Z) + \langle A_{k-1}, Z \rangle \Big\} \ , \tag{A.2}$$

where as before, $V_X(Y) \stackrel{\text{def}}{=} w(Y) - \langle \nabla w(X), Y - X \rangle - w(X)$ is the Bregman divergence of $w(\cdot)$.

Recall that when $w(X) = X \bullet (\log X - I)$ is the entropy regularizer, then $\nabla w(X) = \log X$ and therefore $(\nabla w)^{-1}(A) = e^A$. When $w(X) = -\frac{q}{q-1} \mathrm{Tr} X^{1-1/q}$ is the $\ell_{1-1/q}$ regularizer, then $\nabla w(X) = X^{-1/q}$ and therefore $(\nabla w)^{-1}(A) = A^{-q}$. The rest of the proof holds for both these two types of regularizers.

To compute the minimizer $X_k$ for (A.1), one can take the derivative and demand that $\nabla w(X_k) + \sum_{i=0}^{k-1} A_i - c_k \cdot I = 0$. Here, the extra term $-c_k \cdot I$ comes from the Lagrange multipliers of the linear constraint $\mathrm{Tr}(Z) = I \bullet Z = 1$. (We do not have Lagrange multipliers for the other constraint $Z \succeq 0$ because our gradient $\nabla w(Z)$ is a barrier function and tends to infinite as any eigenvalue of $Z$ tends to zero.) It is now easy to see that $c_k$ is the unique constant that ensures $\sum_{i=0}^{k-1} A_i - c_k I \preceq 0$ (because $\nabla w(X_k) \succeq 0$) and that $\mathrm{Tr} X_k = \mathrm{Tr}\big((\nabla w)^{-1}(c_k I - \sum_{i=0}^{k-1} A_i)\big) = 1$.

To compute the minimizer $X_k$ for (A.2), one can take the derivative and demand that $\nabla w(\widetilde{X}_k) - \nabla w(\widetilde{X}_{k-1}) + A_i - d_k \cdot I = \nabla V_{\widetilde{X}_{k-1}}(\widetilde{X}_k) + A_i - d_k \cdot I = 0$. Here, the extra term $-d_k \cdot I$ again comes from the Lagrange multipliers of the linear constraint $\mathrm{Tr}(Z) = I \bullet Z = 1$. It is now easy to see that $d_k$ is the unique constant that ensures $-\nabla w(\widetilde{X}_{k-1}) + A_i - d_k \cdot I \preceq 0$ (because $\nabla w(\widetilde{X}_k) \succeq 0$) and that $\mathrm{Tr} \widetilde{X}_k = \mathrm{Tr}\big((\nabla w)^{-1}(\nabla w(\widetilde{X}_{k-1}) + d_k I - A_{k-1})\big) = 1$.

To show the equivalence between (A.1) and (A.2), we perform a simple induction. Suppose that $\widetilde{X}_{k-1} = X_{k-1}$, and we wish to prove $\widetilde{X}_k = X_k$.

In this case, we have

$$\begin{aligned} \widetilde{X}_k &= (\nabla w)^{-1}\big(\nabla w(\widetilde{X}_{k-1}) + d_k I - A_{k-1}\big) = (\nabla w)^{-1}\big(\nabla w(X_{k-1}) + d_k I - A_{k-1}\big) \\ &= (\nabla w)^{-1}\Big(c_{k-1} I + d_k I - \sum_{i=0}^{k-1} A_i\Big) \ , \text{ and} \\ X_k &= (\nabla w)^{-1}\Big(c_k I - \sum_{i=0}^{k-1} A_i\Big) \ . \end{aligned}$$

Finally, since $d_k$ is the unique constant that ensures $c_{k-1} I + d_k I - \sum_{i=0}^{k-1} A_i \succeq 0$ and $\mathrm{Tr}\big((\nabla w)^{-1}\big(c_{k-1} I + d_k I - \sum_{i=0}^{k-1} A_i\big)\big) = 1$, while $c_k$ is the unique constant that ensures $c_k I - \sum_{i=0}^{k-1} A_i \succeq 0$ and $\mathrm{Tr}\big((\nabla w)^{-1}\big(c_k I - \sum_{i=0}^{k-1} A_i\big)\big) = 1$, it is obvious to see that $c_k = c_{k-1} + d_k$ and therefore $\widetilde{X}_k = X_k$.

## B Graph Notations

Let $G = (V, E, w)$ be a undirected weighted graph with $n$ vertices and $m$ edges, and each $w_e > 0$ is the weight of edge $e$. Without loss of generality, we study only connected graphs throughout this paper. For every edge $e=(a,b)\in E$, we orient it arbitrarily and denote by $\chi_e \stackrel{\text{def}}{=} \mathbf{e}_a - \mathbf{e}_b \in \mathbb{R}^n$ the characteristic (column) vector of edge $e$.

Let $L_e \stackrel{\text{def}}{=} w_e \chi_e \chi_e^T \in \mathbb{R}^{n\times n}$ be the graph Laplacian of edge $e$, or the edge Laplacian. Let $B\in\mathbb{R}^{m\times n}$ be the incidence matrix where its row corresponding to edge $e$ is the characteristic (row) vector $\chi_e^T$. Define $W=\mathrm{diag}\{w_e\}_{e\in E}$ to be the diagonal matrix of edge weights. The Laplacian with respect to graph $G$ is $L_G \stackrel{\text{def}}{=} B^T W B \in \mathbb{R}^{n\times n}$. It is clear from the definition that $L_G \succeq 0$ is PSD and $L_G = \sum_{e\in E} L_e$. Notice that $\ker(L_G)=\ker(W^{1/2}B)=\mathrm{span}(\mathbb{1})$, and therefore $x^T L_G x = 0$ if and only if $x$ is a constant vector.

Since $L_G$ is symmetric, one can diagonalize it and write $L_G=\sum_{i=1}^{n-1}\lambda_i v_i v_i^T$, where $\lambda_i$'s are the positive eigenvalues of $L_G$ and $v_i$'s are the corresponding set of orthogonal eigenvectors. The Moore-Penrose pseudoinverse of $L_G$ is denoted by $L_G^\dagger \stackrel{\text{def}}{=} \sum_{i=1}^{n-1}\lambda_i^{-1} v_i v_i^T$. For notational convenience, we will stick to $L_G^{-1}$ to denote this pseudoinverse, and often use $L_G^{-2}$ to denote $(L_G^\dagger)^2$, and $L_G^{-1/2}$ to denote $(L_G^\dagger)^{1/2}$, and so on. We remark here that $L_G L_G^{-1} = L_G^{-1} L_G = \sum_i v_i v_i^T = I_{\mathsf{im}(L_G)}$. Here, $I_{\mathsf{im}(L_G)}$ is the identity matrix on the image space of $L_G$, which is just the space spanned by all the vectors orthogonal to $\mathbb{1}$. For notational convenience, we will often abbreviate $I_{\mathsf{im}(L_G)}$ as $I$.

Throughout this paper, whenever related to graph sparsifications, we denote by

$$\widecheck{L}_e \stackrel{\text{def}}{=} L_G^{-1/2} L_e L_G^{-1/2} \quad \text{and} \quad \widehat{L}_e \stackrel{\text{def}}{=} \frac{L_G^{-1/2} L_e L_G^{-1/2}}{L_G^{-1}\bullet L_e} = \frac{\widecheck{L}_e}{L_G^{-1}\bullet L_e} \enspace .$$

Above, $\widecheck{L}_e$ is the *normalized edge Laplacian*, and $\widehat{L}_e$ is the *normalized edge Laplacian scaled by the effective resistance*. ($L_G^{-1}\bullet L_e$ is the "effective resistance" of the edge $e$, see for instance [SS11]). Both of them have rank 1, and it satisfies $\mathrm{Tr}(\widecheck{L}_e)\le 1$ and $\widecheck{L}_e \preceq I$, while $\mathrm{Tr}(\widehat{L}_e)=1$ and $\widehat{L}_e \preceq I$.

It is easy to check from the above definition that $\sum_e \widecheck{L}_e = I_{\mathsf{im}(L_G)}$. In addition, letting $w_e = L_G^{-1}\bullet L_e$ be the effective resistence of edge $e$, then $\sum_e w_e \widehat{L}_e = I_{\mathsf{im}(L_G)}$ as well. Notice that $\sum_e w_e = \mathrm{Tr} I_{\mathsf{im}(L_G)} = n-1$, the dimension of $I_{\mathsf{im}(L_G)}$ (see [SS11]).

**From Graph Sparsification to Rank-1 Decomposition Sparsification.** As originally shown in [BSS14], one can easily translate the problem of graph spectral sparsification (see Theorem 1) into that of sparsifying sums of rank-1 matrices (see Theorem 2). Indeed, because $I_{\mathsf{im}(L_G)} = \sum_{e\in[m]}\widecheck{L}_e$ is a summation of rank-1 matrices, if one can find scalars $s_e\ge 0$ (as per Theorem 2) that satisfies $I_{\mathsf{im}(L_G)} \preceq \sum_{e\in[m]} s_e \widecheck{L}_e \preceq (1+\varepsilon) I_{\mathsf{im}(L_G)}$, this immediately implies, by the definition of $\widecheck{L}_e$, that $L_G \preceq \sum_{e\in[m]} s_e L_e \preceq (1+\varepsilon) L_G$.

# C Weak Unweighted Sparsifier

In this section, we consider the weak *unweighted spectral sparsification* problem very recently studied by Anderson, Gu and Melgaard [AGM14]: for any value $\kappa\in[1,m/n]$, find a $\kappa$-spectral sparsifier of $G$ containing $O(m/\kappa)$ distinct edges from $E$, *without reweighting.* We show that our regret minimization framework allows us to design a simple and almost-quadratic-time algorithm for this problem, improving from the quartic time complexity of [AGM14].

Formally, given any weighted undirected graph $G=(V,E,w)$ with $n$ vertices and $m$ edges, and any value $\kappa\in[1,m/n]$, the task it to find a subset $E_0\subseteq E$ containing $O(m/\kappa)$ distinct edges such that

$$\frac{1}{\kappa} L_G \preceq \sum_{e\in E_0} L_e \preceq L_G \enspace .$$

This is an unweighted sparsification problem because one is not allowed to reweight the edges in $E_0$, in contrast to Theorem 1; and we call it a weak sparsifier because $\kappa$ is usually large.

Similar to Appendix B, one can easily reduce this graph sparsification problem to sparsifying sums of rank-1 matrices. Given $m$ rank-1 PSD matrices $\check{L}_1, \dots, \check{L}_m \in \mathbb{R}^{n\times n}$ that satisfies $I = \sum_{e\in[m]} \check{L}_e$, and given some $\kappa \in [1, m/n]$, find a subset $E_0 \subseteq [m]$ with $O(m/\kappa)$ distinct elements satisfying $\sum_{e\in E_0} \check{L}_e \succeq \frac{1}{\kappa} I$.

*(In this section, one should feel free to coincide this $\check{L}_e$ with the 'normalized edge Laplacian' introduced in Section B; but $\widehat{L}_e$ needs not coincide with any graph Laplacian in general.)*

We solve this weak unweighted sparsification problem via the following reduction to regret minimization.

If $\kappa \leq 9$, we output $E_0 = E$ and are done. Otherwise, we choose the $\ell_{1/2}$ regularizer and parameter $\alpha = 4\sqrt{n}\kappa$ for $\texttt{MirrorDescent}_{\ell_{1/2}}$. At each iteration $k = 0, 1, \dots, T-1$, we define $e_k = e$ to be the index $e \in [m]$ that maximizes the quantity $\frac{X_k \bullet \check{L}_e}{1 + X_k^{1/2} \bullet \alpha \check{L}_e}$ among all edges not chosen before —i.e., all edges in $E \setminus \{e_0, e_1, \dots, e_{k-1}\}$. Next, we feed $F_k = \check{L}_{e_k}$ as the feedback matrix to $\texttt{MirrorDescent}_{\ell_{1/2}}$, and compute $X_{k+1}$ of the next iteration.

Let us now state a simple property for the selected matrix $\check{L}_{e_k}$ using an averaging argument:

**Claim C.1.** *For each $k = 0, 1, \dots, T-1$, we either have $\sum_{j=0}^{k-1} \check{L}_{e_j} \succeq \frac{1}{\kappa} I$ or $\frac{X_k \bullet \check{L}_{e_k}}{1 + X_k^{1/2} \bullet \alpha \check{L}_{e_k}} \geq \frac{1}{6m}$.*

*Proof.* Let us recall that by the definition of $\texttt{MirrorDescent}_{\ell_{1/2}}$, we have

$$X_k = \Big(\alpha \sum_{j=0}^{k-1} \check{L}_{e_j} - c_k I\Big)^{-2} ,$$

where $c_k > 0$ is the unique constant that makes $\alpha \sum_{j=0}^{k-1} \check{L}_{e_j} \succ c_k I$ and $\mathrm{Tr} X_k = 1$. Note that if $c_k/\alpha \geq \frac{1}{\kappa}$ then we already have $\sum_{j=0}^{k-1} \check{L}_{e_j} \succ \frac{c_k}{\alpha} I \succeq \frac{1}{\kappa} I$. Therefore, we can assume $c_k/\alpha < \frac{1}{\kappa}$ for the rest of the proof.

One one hand, we have

$$\sum_{e\notin\{e_0,\dots,e_{k-1}\}} X_k \bullet \check{L}_e = X_k \bullet \Big(I - \sum_{j=0}^{k-1} \check{L}_{e_j}\Big) = X_k \bullet \Big(I - \frac{c_k}{\alpha} I - \frac{X_k^{-1/2}}{\alpha}\Big)$$

$$= (1 - \frac{c_k}{\alpha}) - \frac{\mathrm{Tr} X_k^{1/2}}{\alpha} > 1 - \frac{1}{\kappa} - \frac{\sqrt{n}}{\alpha} > \frac{5}{6} , \tag{C.1}$$

where the first inequality is due to $\mathrm{Tr} X_k^{1/2} \leq \sqrt{n}$ and the second inequality is due to our choice of $\alpha = 4\sqrt{n}\kappa$ and the fact that $\kappa > 9$.

On the other hand, we have

$$\sum_{e\notin\{e_0,\dots,e_{k-1}\}} \frac{1}{6m}\big(1 + X_k^{1/2} \bullet \alpha \check{L}_e\big) \leq \frac{1}{6} + \frac{\alpha}{6m} X_k^{1/2} \bullet \sum_{e\notin\{e_0,\dots,e_{k-1}\}} \alpha \check{L}_e$$

$$\leq \frac{1}{6} + \frac{\alpha}{6m} X_k^{1/2} \bullet I \leq \frac{1}{6} + \frac{\alpha\sqrt{n}}{6m} = \frac{1}{6} + \frac{4n\kappa}{6m} \leq \frac{5}{6} , \tag{C.2}$$

where the second inequality is because $\sum_{e\notin\{e_0,\dots,e_{k-1}\}} \check{L}_e \preceq \sum_{e\in[m]} \check{L}_e = I$, the third inequality is because $\mathrm{Tr} X_k^{1/2} \leq \sqrt{n}$, and the fourth inequality is because $\kappa \leq m/n$.

Combining (C.1) and (C.2), we conclude that there exists at least some index $e \in [m] \setminus \{e_0, \dots, e_{k-1}\}$ satisfying that $X_k \bullet \check{L}_e \geq \frac{1}{7m}\big(1 + X_k^{1/2} \bullet \alpha \check{L}_e\big)$, finishing the proof of the claim. $\square$

Now we are ready to apply Theorem 3.2, the regret bound, with our choice of $F_k = \check{L}_{e_k}$:

$$
\begin{aligned}
\forall U \in \Delta_{n\times n}, \qquad \sum_{k=0}^{T-1}\langle \check{L}_{e_k}, U\rangle &\geq \sum_{k=0}^{T-1}\langle \check{L}_{e_k}, X_k\rangle - \alpha\frac{\mathrm{Tr}(X_k\check{L}_{e_k}X_k^{1/2}\check{L}_{e_k})}{1+X_k^{1/2}\bullet\alpha\check{L}_{e_k}} - \frac{2\sqrt{n}}{\alpha} \\
&= \sum_{k=0}^{T-1}\langle \check{L}_{e_k}, X_k\rangle\Big(1 - \frac{X_k^{1/2}\bullet\alpha\check{L}_{e_k}}{1+X_k^{1/2}\bullet\alpha\check{L}_{e_k}}\Big) - \frac{2\sqrt{n}}{\alpha} \\
&= \sum_{k=0}^{T-1}\frac{\check{L}_{e_k}\bullet X_k}{1+X_k^{1/2}\bullet\alpha\check{L}_{e_k}} - \frac{2\sqrt{n}}{\alpha} \ . \qquad (C.3)
\end{aligned}
$$

We will now choose $T = 9m/\kappa$. (Notice that $T < m$ because $\kappa > 9$.) There are two possibilities according to Claim C.1.

In the first case, we have $\sum_{j=0}^{k-1}\check{L}_{e_j} \succeq \frac{1}{\kappa}I$ for some $k = 0, 1, \dots, T-1$ and we are done: that is, defining $E_0 \stackrel{\text{def}}{=} \{e_0, e_1, \dots, e_{k-1}\}$, we have that $|E_0| \leq T = O(m/\kappa)$ and $I \succeq \sum_{e\in E_0}\check{L}_e \succeq \frac{1}{\kappa}I$.

In the second case, we have $\frac{X_k\bullet\check{L}_{e_k}}{1+X_k^{1/2}\bullet\alpha\check{L}_{e_k}} \geq \frac{1}{6m}$ for all $k = 0, 1, \dots, T-1$. Substituting this into (C.3), and choosing $U$ to be the rank 1 matrix corresponding to the smallest eigenvalue of $\sum_{k=0}^{T-1}\check{L}_{e_k}$, we conclude that

$$
\lambda_{\mathsf{min}}\Big(\sum_{k=0}^{T-1}\check{L}_{e_k}\Big) \geq \sum_{k=0}^{T-1}\frac{1}{6m} - \frac{1}{2\kappa} = \frac{1}{\kappa} \ .
$$

Therefore, defining $E_0 \stackrel{\text{def}}{=} \{e_0, e_1, \dots, e_{T-1}\}$, we also have $|E_0| = T = O(m/\kappa)$ and $I \succeq \sum_{e\in E_0}\check{L}_e \succeq \frac{1}{\kappa}I$. In sum,

**Theorem C.2.** *Given a decomposition $I = \sum_{e\in[m]}\check{L}_e$ of rank-1 PSD matrices, and given some $\kappa \in [1, m/n]$, the above algorithm finds a subset $E_0 \subseteq [m]$ with $O(\frac{m}{\kappa})$ distinct elements satisfying $I \succeq \sum_{e\in E_0}\check{L}_e \succeq \frac{1}{\kappa}I$.*

We remark here that for graph sparsification, the above algorithm can be implemented to run in time $\widetilde{O}(m^{3/2}n)$, and can be improved to $\widetilde{O}(m^{1+1/q}n)$ for any even integer constant $q \geq 2$ if the $\ell_{1-1/q}$ regularizer is used instead of $\ell_{1/2}$. We ignore the implementation details in this version of the paper because it is very similar to the details discussed in Section 6.

# D Proof of Lemma 2.1

We state some classical properties for Bregman divergence, which are classical and can be found in for instance [CL06].

**Lemma 2.1.** *The following properties hold for Bregman divergence.*

- *Non-negativity: $V_X(Y) \geq 0$ for all $X, Y \geq 0$.*
- *The "three-point equality": $\langle \nabla w(X) - \nabla w(Y), X - U\rangle = V_X(U) - V_Y(U) + V_Y(X)$.*
- *Given $\widetilde{X} \succeq 0$ and $X = \arg\min_{Z\in\Delta_{n\times n}} V_{\widetilde{X}}(Z)$ as the* Bregman projection*, we have the "generalized Pythagorean theorem" for all $U \in \Delta$: $V_{\widetilde{X}}(U) \geq V_X(U) + V_{\widetilde{X}}(X) \geq V_X(U)$.*

*Proof.* The non-negativity follows by definition from the convexity of $w(X)$. For every $U \succeq 0$, the "three-point equality" follows from the following inequality.

$$
\begin{aligned}
\langle \nabla w(Y) - \nabla w(Y), Y - U\rangle &= (w(U) - w(Y) - \langle \nabla w(Y), U - Y\rangle) - (w(U) - w(Y) - \langle w(Y), U - Y\rangle) \\
&\quad - (w(Y) - w(Y) - \langle \nabla w(Y), Y - Y\rangle) \\
&= V_Y(U) - V_Y(U) - V_Y(Y) \ .
\end{aligned}
$$

For the generalized Pythagorean theorem, we only need to prove $V_{\widetilde{X}}(U) \geq V_X(U) + V_{\widetilde{X}}(X)$ because the second inequality follows from the non-negativity of $V_{\widetilde{X}}(X)$. To provide the simplest proof, we only focus on the special case when $w(X) = -\frac{q}{q-1}\mathrm{Tr}X^{1-1/q}$. (The proof for the entropy regularizer is similar, while the proof for the most general Legendre function case is more involved.)

By definition,

$$\begin{aligned} V_X(U) + V_{\widetilde{X}}(X) &= X^{-1/q} \bullet U + \frac{1}{q-1}\mathrm{Tr}X^{1-1/q} - \frac{q}{q-1}\mathrm{Tr}U^{1-1/q} \\ &\quad + \widetilde{X}^{-1/q} \bullet X + \frac{1}{q-1}\mathrm{Tr}\widetilde{X}^{1-1/q} - \frac{q}{q-1}\mathrm{Tr}X^{1-1/q} \\ V_{\widetilde{X}}(U) &= \widetilde{X}^{-1/q} \bullet U + \frac{1}{q-1}\mathrm{Tr}\widetilde{X}^{1-1/q} - \frac{q}{q-1}\mathrm{Tr}U^{1-1/q} \enspace . \end{aligned}$$

Therefore,

$$\begin{aligned} V_{\widetilde{X}}(U) - (V_X(U) + V_{\widetilde{X}}(X)) &= \widetilde{X}^{-1/q} \bullet U - X^{-1/q} \bullet U - \widetilde{X}^{-1/q} \bullet X + \mathrm{Tr}X^{1-1/q} \\ &= (\widetilde{X}^{-1/q} - X^{-1/q}) \bullet (U - X) \enspace . \end{aligned}$$

Since $V_{\widetilde{X}}(U)$ is a convex function and $X = \arg\min_{z\in\Delta} V_{\widetilde{X}}(z)$, for any $U \in \Delta_{n\times n}$ we must have

$$\langle \nabla V_{\widetilde{X}}(X), U - X\rangle \geq 0 \Longleftrightarrow \langle -X^{-1/q} + \widetilde{X}^{-1/q}, U - X\rangle \geq 0 \enspace .$$

This concludes the proof of the lemma. □

# E Missing Proofs in Section 3

**Theorem 3.1.** *In* $\mathtt{MirrorDescent}_{\mathsf{exp}}$*, if the parameter* $\alpha > 0$ *satisfies* $\alpha F_k \succeq -I$ *for all iterations* $k = 0, 1, \ldots, T-1$*, then, for every* $U \in \Delta_{n\times n}$*,*

$$R(U) \stackrel{\mathrm{def}}{=} \sum_{k=0}^{T-1} \langle F_k, X_k - U\rangle \leq \alpha \sum_{k=0}^{T-1} \big(X_k \bullet |F_k|\big) \cdot \|F_k\|_{\mathsf{spe}} + \frac{V_{X_0}(U)}{\alpha} \enspace .$$

*We note that* $V_{X_0}(U) \leq \log n$*.*

*Proof.* We prove the theorem by using a two-step description of the mirror descent. For every $k \geq 0$, define $\widetilde{X}_{k+1} \stackrel{\mathrm{def}}{=} \arg\min_{Z\succeq 0}\{V_{X_k}(Z) + \alpha\langle F_k, Z\rangle\}$, where the minimization is over all $Z \succeq 0$, rather than $Z \in \Delta_{n\times n}$. This minimizer $\widetilde{X}_{k+1}$ certainly exists (and equals to $\exp^{\log X_k - \alpha F_k}$), and it is not hard to verify that $X_{k+1} = \arg\min_{Z\in\Delta_{n\times n}}\{V_{\widetilde{X}_{k+1}}(Z)\}$. In other words, one can describe the update $X_k \to X_{k+1}$ by adding an intermediate stage $X_k \to \widetilde{X}_{k+1} \to X_{k+1}$. We also assume that initially we have $\widetilde{X}_0 \stackrel{\mathrm{def}}{=} X_0$.

Noticing that the definition of $\widetilde{X}_{k+1}$ implies that $\nabla V_{X_k}(\widetilde{X}_{k+1}) + \alpha F_k = 0$, which by the definition of $V_X(Y)$ is equivalent to $\nabla w(X_k) - \nabla w(\widetilde{X}_{k+1}) = \alpha F_k$. Therefore,

$$\begin{aligned} \langle \alpha F_k, X_k - U\rangle = \langle \nabla w(X_k) - \nabla w(\widetilde{X}_{k+1}), X_k - U\rangle &= V_{X_k}(U) - V_{\widetilde{X}_{k+1}}(U) + V_{\widetilde{X}_{k+1}}(X_k) \\ &\leq V_{\widetilde{X}_k}(U) - V_{\widetilde{X}_{k+1}}(U) + V_{\widetilde{X}_{k+1}}(X_k) \enspace . \end{aligned} \tag{E.1}$$

Above, the second equality is due to the three-point equality and the only inequality is due to the generalized Pythagorean theorem of Bregman divergence (see Lemma 2.1). Now,

$$\begin{aligned} V_{\widetilde{X}_{k+1}}(X_k) &= X_k \bullet (\log X_k - \log \widetilde{X}_{k+1}) + \mathrm{Tr}\widetilde{X}_{k+1} - \mathrm{Tr}X_k \\ &= X_k \bullet \alpha F_k + \mathrm{Tr}\big(e^{\log X_k - \alpha F_k}\big) - \mathrm{Tr}X_k \stackrel{①}{\leq} X_k \bullet \alpha F_k + X_k \bullet e^{-\alpha F_k} - \mathrm{Tr}X_k \\ &\stackrel{②}{\leq} X_k \bullet \alpha F_k + X_k \bullet (I - \alpha F_k + \alpha^2 F_k^2) - \mathrm{Tr}X_k = \alpha^2 \cdot X_k \bullet F_k^2 \stackrel{③}{\leq} \alpha^2 \cdot (X_k \bullet |F_k|)\|F_k\|_{\mathsf{spe}} \enspace . \end{aligned}$$

Above, ① is due to the Golden-Thompson inequality. ② follows because $e^{-\alpha A} \preceq I - \alpha A + \alpha^2 A^2$, which can be proved after transforming into its eigenbasis, and then using the fact that $e^{-a} \leq 1 - a + a^2$ for all $a \geq -1$. ③ follows because $F_k^2 \preceq \|F_k\|_{\mathsf{spe}} \cdot |F_k|$.

Finally, substituting the above upper bound into (E.1) and telescoping it for $k = 0, \dots, T-1$, we obtain

$$\sum_{k=0}^{T-1} \langle F_k, X_k - U \rangle \leq \frac{V_{\widetilde{X}_0}(U) - V_{\widetilde{X}_T}(U)}{\alpha} + \alpha \sum_{k=0}^{T-1} (X_k \bullet |F_k|) \|F_k\|_{\mathsf{spe}} \ .$$

The desired result of this theorem now follows from the above inequality and the simple upper bound $V_{\widetilde{X}_0}(U) = V_{X_0}(U) \leq \log n$ and the nonnegativity $V_{\widetilde{X}_T}(U) \geq 0$. □

**Theorem 3.2.** *In* $\mathtt{MirrorDescent}_{\ell_{1/2}}$*, if the parameter* $\alpha > 0$*, and the loss matrix* $F_k$ *is rank one and satisfies* $X_k^{1/2} \bullet \alpha F_k > -1$ *for all* $k$*, then, for every* $U \in \Delta_{n \times n}$*,*

$$R(U) \stackrel{\text{def}}{=} \sum_{k=0}^{T-1} \langle F_k, X_k - U \rangle \leq \alpha \cdot \sum_{k=0}^{T-1} \frac{(X_k \bullet F_k)(X_k^{1/2} \bullet F_k)}{1 + X_k^{1/2} \bullet \alpha F_k} + \frac{V_{X_0}(U)}{\alpha} \ .$$

*If we instead have* $X_k^{1/2} \bullet \alpha F_k \geq -\frac{1}{2}$*, the above bound can be simplified as*

$$R(U) \stackrel{\text{def}}{=} \sum_{k=0}^{T-1} \langle F_k, X_k - U \rangle \leq 2\alpha \cdot \sum_{k=0}^{T-1} (X_k \bullet F_k)(X_k^{1/2} \bullet F_k) + \frac{V_{X_0}(U)}{\alpha} \ .$$

*We note that* $V_{X_0}(U) \leq 2\sqrt{n}$*.*

*Proof.* We prove the theorem by using a two-step description of the mirror descent. For every $k \geq 0$, define $\widetilde{X}_{k+1} \stackrel{\text{def}}{=} \arg\min_{Z \succeq 0} \{V_{X_k}(Z) + \alpha \langle F_k, Z \rangle\}$, where the minimization is over all $Z \succeq 0$, rather than $Z \in \Delta_{n \times n}$. We claim that this minimizer $\widetilde{X}_{k+1}$ exists and is strictly positive definite, because one can choose $Z = \widetilde{X}_{k+1} = (X_k^{-1/2} + \alpha F_k)^{-2} \succ 0$ to make the gradient zero:

$$\nabla V_{X_k}\big(\widetilde{X}_{k+1}\big) + \alpha F_k = \nabla w(\widetilde{X}_{k+1}) - \nabla w(X_k) + \alpha F_k = -\widetilde{X}_{k+1}^{-1/2} + X_k^{-1/2} + \alpha F_k = 0 \ . \tag{E.2}$$

This uses our assumption $X_k^{1/2} \bullet \alpha F_k > -1$ which is equivalent to $\alpha F_k \succ -X_k^{-1/2}$,[8] so as to ensure that $\widetilde{X}_{k+1}$ is well defined.

Next, it is easy to verify that $X_{k+1} = \arg\min_{Z \in \Delta_{n \times n}} \{V_{\widetilde{X}_{k+1}}(Z)\}$. In other words, one can describe the update $X_k \to X_{k+1}$ by adding an intermediate stage $X_k \to \widetilde{X}_{k+1} \to X_{k+1}$. We assume for notational simplicity that $\widetilde{X}_0 \stackrel{\text{def}}{=} X_0$.

Using (E.2), we easily obtain that

$$\begin{aligned} \langle \alpha F_k, X_k - U \rangle = \langle \nabla w(X_k) - \nabla w(\widetilde{X}_{k+1}), X_k - U \rangle &= V_{X_k}(U) - V_{\widetilde{X}_{k+1}}(U) + V_{\widetilde{X}_{k+1}}(X_k) \\ &\leq V_{\widetilde{X}_k}(U) - V_{\widetilde{X}_{k+1}}(U) + V_{\widetilde{X}_{k+1}}(X_k) \ . \end{aligned} \tag{E.3}$$

Above, the second equality is due to the three-point equality and the only inequality is due to the generalized Pythagorean theorem of Bregman divergence (see Lemma 2.1).

We now exactly compute $V_{\widetilde{X}_{k+1}}(X_k)$ in two cases.

- If $\alpha F_k = -uu^T$ is negative semidefinite, using the Sherman-Morrison formula,

$$\mathrm{Tr}\widetilde{X}_{k+1}^{1/2} = \mathrm{Tr}\big((X_k^{-1/2} - uu^T)^{-1}\big) = \mathrm{Tr}\Big(X_k^{1/2} + \frac{X_k^{1/2} uu^T X_k^{1/2}}{1 - u^T X_k^{1/2} u}\Big) \ .$$

[8] This is because, if $F_k = -uu^T$, then $X_k^{1/2} \bullet (-\alpha uu^T) > -1$ is equivalent to $\alpha u^T X_k^{1/2} u < 1$, which is further equivalent to $\alpha \mathrm{Tr} X_k^{1/4} uu^T X_k^{1/4} < 1$. However, since $X_k^{1/4} uu^T X_k^{1/4}$ is a rank-1 matrix, this is finally equivalent to $\alpha uu^T \prec X_k^{-1/2}$.

Therefore,

$$V_{\widetilde{X}_{k+1}}(X_k) = \widetilde{X}_{k+1}^{-1/2} \bullet X_k + \mathrm{Tr}\widetilde{X}_{k+1}^{1/2} - 2\mathrm{Tr}X_k^{1/2} = (X_k^{-1/2} - uu^T) \bullet X_k + \mathrm{Tr}\widetilde{X}_{k+1}^{1/2} - 2\mathrm{Tr}X_k^{1/2}$$
$$= -uu^T \bullet X_k + \big(\mathrm{Tr}\widetilde{X}_{k+1}^{1/2} - \mathrm{Tr}X_k^{1/2}\big) = -u^T X_k u + \frac{u^T X_k u}{1 - u^T X_k^{1/2} u}$$
$$= \frac{u^T X_k u \cdot u^T X_k^{1/2} u}{1 - u^T X_k^{1/2} u} = \alpha^2 \frac{(X_k \bullet F_k)(X_k^{1/2} \bullet F_k)}{1 + X_k^{1/2} \bullet \alpha F_k} \ .$$

- If $\alpha F_k = uu^T$ is positive semidefinite, using the Sherman-Morrison formula,

$$\mathrm{Tr}\widetilde{X}_{k+1}^{1/2} = \mathrm{Tr}\big((X_k^{-1/2} + uu^T)^{-1}\big) = \mathrm{Tr}\Big(X_k^{1/2} - \frac{X_k^{1/2} uu^T X_k^{1/2}}{1 + u^T X_k^{1/2} u}\Big) \ .$$

Therefore,

$$V_{\widetilde{X}_{k+1}}(X_k) = \widetilde{X}_{k+1}^{-1/2} \bullet X_k + \mathrm{Tr}\widetilde{X}_{k+1}^{1/2} - 2\mathrm{Tr}X_k^{1/2} = (X_k^{-1/2} + uu^T) \bullet X_k + \mathrm{Tr}\widetilde{X}_{k+1}^{1/2} - 2\mathrm{Tr}X_k^{1/2}$$
$$= uu^T \bullet X_k + \big(\mathrm{Tr}\widetilde{X}_{k+1}^{1/2} - \mathrm{Tr}X_k^{1/2}\big) = u^T X_k u + \frac{u^T X_k u}{1 + u^T X_k^{1/2} u}$$
$$= \frac{u^T X_k u \cdot u^T X_k^{1/2} u}{1 + u^T X_k^{1/2} u} = \alpha^2 \frac{(X_k \bullet F_k)(X_k^{1/2} \bullet F_k)}{1 + X_k^{1/2} \bullet \alpha F_k} \ .$$

Finally, substituting the above computation of $V_{\widetilde{X}_{k+1}}(X_k)$ into (E.3) and telescoping it for $k = 0, \dots, T-1$, we obtain

$$\sum_{k=0}^{T-1} \langle F_k, X_k - U \rangle \le \frac{V_{\widetilde{X}_0}(U) - V_{\widetilde{X}_T}(U)}{\alpha} + \alpha \sum_{k=0}^{T-1} \frac{(X_k \bullet F_k)(X_k^{1/2} \bullet F_k)}{1 + X_k^{1/2} \bullet \alpha F_k} \ .$$

The desired result of this theorem now follows from the above inequality and the simple upper bound $V_{\widetilde{X}_0}(U) = V_{X_0}(U) \le 2\sqrt{n}$ and the nonnegativity $V_{\widetilde{X}_T}(U) \ge 0$. □

The next theorem generalizes Theorem 3.2 to high rank loss matrices and $\ell_{1-1/q}$-regularizers with $q \ge 2$. The key idea is to replace the use of the Sherman-Morrison formula in the proof of Theorem 3.2 with the Woodbury formula so as to allow $F_k$ to be of high rank. It also uses the Lieb-Thirring trace inequality to handle arbitrary $q \ge 2$.

**Theorem 3.3.** *In* $\mathtt{MirrorDescent}_{\ell_{1-1/q}}$ *with* $q \ge 2$ *and* $\alpha > 0$*, if the loss matrix* $F_k$ *is either positive or negative semidefinite and satisfies* $\alpha X_k^{1/2q} F_k X_k^{1/2q} \succeq -\frac{1}{2q} I$ *for all* $k$*, then,*

$$\forall U \in \Delta_{n\times n}, \quad R(U) \stackrel{\mathrm{def}}{=} \sum_{k=0}^{T-1} \langle F_k, X_k - U \rangle \le O(q\alpha) \sum_{k=0}^{T-1} (X_k \bullet |F_k|) \cdot \|X_k^{1/2q} F_k X_k^{1/2q}\|_{\mathsf{spe}} + \frac{V_{X_0}(U)}{\alpha} \ .$$

*We note that* $V_{X_0}(U) \le \frac{q}{q-1} n^{1/q}$.

*Proof.* We prove the theorem by using a two-step description of the mirror descent. For every $k \ge 0$, define $\widetilde{X}_{k+1} \stackrel{\mathrm{def}}{=} \arg\min_{Z \succeq 0} \{V_{X_k}(Z) + \alpha \langle F_k, Z \rangle\}$, where the minimization is over all $Z \succeq 0$, rather than $Z \in \Delta_{n\times n}$. We claim that this minimizer $\widetilde{X}_{k+1}$ exists and is strictly positive definite, because one can choose $Z = \widetilde{X}_{k+1} = (X_k^{-1/q} + \alpha F_k)^{-q} \succ 0$ to make the gradient zero:

$$\nabla V_{X_k}\big(\widetilde{X}_{k+1}\big) + \alpha F_k = \nabla w(\widetilde{X}_{k+1}) - \nabla w(X_k) + \alpha F_k = -\widetilde{X}_{k+1}^{-1/q} + X_k^{-1/q} + \alpha F_k = 0 \ . \quad \text{(E.4)}$$

This uses our assumption $\alpha X_k^{1/2q} F_k X_k^{1/2q} \succeq -\frac{1}{2q} I$ which certainly implies $\alpha F_{k,i} \succeq -\frac{1}{2} X_k^{-1/q}$, so as to ensure that $\widetilde{X}_{k+1}$ is well defined.

Next, it is easy to verify that $X_{k+1} = \arg\min_{Z \in \Delta}\{V_{\widetilde{X}_{k+1}}(Z)\}$. In other words, one can describe the update $X_k \to X_{k+1}$ by adding an intermediate stage $X_k \to \widetilde{X}_{k+1} \to X_{k+1}$. We assume for notational simplicity that $\widetilde{X}_0 \stackrel{\text{def}}{=} X_0$.

Using (E.4), we easily obtain that

$$\begin{aligned}\langle \alpha F_k, X_k - U\rangle = \langle \nabla w(X_k) - \nabla w(\widetilde{X}_{k+1}), X_k - U\rangle &= V_{X_k}(U) - V_{\widetilde{X}_{k+1}}(U) + V_{\widetilde{X}_{k+1}}(X_k) \\ &\leq V_{\widetilde{X}_k}(U) - V_{\widetilde{X}_{k+1}}(U) + V_{\widetilde{X}_{k+1}}(X_k) \ . \qquad \text{(E.5)}\end{aligned}$$

Above, the second equality is due to the three-point equality and the only inequality is due to the generalized Pythagorean theorem of Bregman divergence (see Lemma 2.1).

We now upper bound $V_{\widetilde{X}_{k+1}}(X_k)$ in two cases: the case when $\alpha F_k = -PP^T \preceq 0$ and the case when $\alpha F_k = PP^T \succeq 0$. In both cases, we denote by $\beta \stackrel{\text{def}}{=} \alpha \|X_k^{1/2q} F_k X_k^{1/2q}\|_{\mathsf{spe}} = \|X_k^{1/2q} PP^T X_k^{1/2q}\|_{\mathsf{spe}}$. Notice that this implies [9]

$$X_k^{1/2q} PP^T X_k^{1/2q} \preceq \beta I \quad \text{and} \quad P^T X_k^{1/q} P \preceq \beta I \ . \qquad \text{(E.6)}$$

- If $\alpha F_k = -PP^T$, we have $X_k^{-1/q} \succ PP^T$ and $\beta \leq \frac{1}{2q}$ by our assumption, so using the Sherman-Morrison-Woodbury formula,

$$\begin{aligned}\mathrm{Tr}\widetilde{X}_{k+1}^{1-1/q} = \mathrm{Tr}\big((X_k^{-1/q} - PP^T)^{-1}\big)^{q-1} &= \mathrm{Tr}\Big(X_k^{1/q} + X_k^{1/q} P\big(I - P^T X_k^{1/q} P\big)^{-1} P^T X_k^{1/q}\Big)^{q-1} \\ &\leq \mathrm{Tr}\Big(X_k^{1/q} + \frac{X_k^{1/q} PP^T X_k^{1/q}}{1-\beta}\Big)^{q-1} \ ,\end{aligned}$$

  where the last inequality follows because $(I - P^T X_k^{1/q} P)^{-1} \preceq \frac{1}{1-\beta} I$ owing to (E.6), as well as $A \preceq B \implies \mathrm{Tr}A^n \leq \mathrm{Tr}B^n$. We continue and write

$$\begin{aligned}\mathrm{Tr}\widetilde{X}_{k+1}^{1-1/q} \leq \mathrm{Tr}\Big(X_k^{1/q} + \frac{X_k^{1/q} PP^T X_k^{1/q}}{1-\beta}\Big)^{q-1} &= \mathrm{Tr}\Big(X_k^{1/2q}\big(I + \frac{X_k^{1/2q} PP^T X_k^{1/2q}}{1-\beta}\big) X_k^{1/2q}\Big)^{q-1} \\ &\leq \mathrm{Tr}\Big(X_k^{(q-1)/2q}\big(I + \frac{X_k^{1/2q} PP^T X_k^{1/2q}}{1-\beta}\big)^{q-1} X_k^{(q-1)/2q}\Big) \\ &= \mathrm{Tr}\Big(X_k^{1-1/q}\big(I + \frac{X_k^{1/2q} PP^T X_k^{1/2q}}{1-\beta}\big)^{q-1}\Big) \ ,\end{aligned}$$

  where the inequality uses the Lieb-Thirring trace inequality (which relies on the fact that $q-1 \geq 1$). Finally, denoting by $D \stackrel{\text{def}}{=} \frac{X_k^{1/2q} PP^T X_k^{1/2q}}{1-\beta} \preceq \frac{\beta}{1-\beta} I$ (which uses (E.6) again), we have

$$\big(I + D\big)^{q-1} \preceq I + (q-1)D + O(q^2\beta) \cdot D \ .$$

  This matrix inequality can be proved by first turning into its eigenbasis, and then verifying that $(1+x)^{q-1} \leq 1 + (q-1)x + O(q^2\beta)x$ for all $x \in [0, \frac{\beta}{1-\beta}]$ (which uses the fact that $\beta \leq 1/2q$).

[9] The second inequality is because $P^T X_k^{1/q} P = (P^T X_k^{1/2q})(P^T X_k^{1/2q})^T$ and has the same largest eigenvalue as $(P^T X_k^{1/2q})^T (P^T X_k^{1/2q}) = X_k^{1/2q} PP^T X_k^{1/2q}$.

Using this inequality, we conclude that

$$\begin{aligned}
\mathrm{Tr}\widetilde{X}_{k+1}^{1-1/q} &\le \mathrm{Tr}\Big(X_k^{1-1/q}\big(I+\frac{X_k^{1/2q}PP^TX_k^{1/2q}}{1-\beta}\big)^{q-1}\Big)\\
&\le \mathrm{Tr}\Big(X_k^{1-1/q}\Big(I+\Big((q-1)+O(q^2\beta)\Big)\frac{X_k^{1/2q}PP^TX_k^{1/2q}}{1-\beta}\Big)\Big)\\
&= \mathrm{Tr}X_k^{1-1/q} + (q-1)(1+O(q\beta))X_k\bullet PP^T \enspace.
\end{aligned}$$

Therefore,

$$\begin{aligned}
V_{\widetilde{X}_{k+1}}(X_k) &= \widetilde{X}_{k+1}^{-1/q}\bullet X_k + \frac{1}{q-1}\mathrm{Tr}\widetilde{X}_{k+1}^{1-1/q} - \frac{q}{q-1}\mathrm{Tr}X_k^{1-1/q}\\
&= (X_k^{-1/q}-PP^T)\bullet X_k + \frac{1}{q-1}\mathrm{Tr}\widetilde{X}_{k+1}^{1-1/q} - \frac{q}{q-1}\mathrm{Tr}X_k^{1-1/q}\\
&= -PP^T\bullet X_k + \frac{1}{q-1}\big(\mathrm{Tr}\widetilde{X}_{k+1}^{1-1/q}-\mathrm{Tr}X_k^{1-1/q}\big)\\
&= O(q\beta)\cdot PP^T\bullet X_k = O(q\alpha^2)(X_k\bullet|F_k|)\cdot\|X_k^{1/2q}F_kX_k^{1/2q}\|_{\mathsf{spe}} \enspace.
\end{aligned}$$

- If $\alpha F_k = PP^T$, using the Sherman-Morrison-Woodbury formula,

$$\begin{aligned}
\mathrm{Tr}\widetilde{X}_{k+1}^{1-1/q} = \mathrm{Tr}\big((X_k^{-1/q}+PP^T)^{-1}\big)^{q-1} &= \mathrm{Tr}\Big(X_k^{1/q}-X_k^{1/q}P(I+P^TX_k^{1/q}P)^{-1}P^TX_k^{1/q}\Big)^{q-1}\\
&\le \mathrm{Tr}\Big(X_k^{1/q}-\frac{X_k^{1/q}PP^TX_k^{1/q}}{1+\beta}\Big)^{q-1} \enspace,
\end{aligned}$$

where the last inequality follows because $(I+P^TX_k^{1/q}P)^{-1}\succeq\frac{1}{1+\beta}I$ owing to (E.6), as well as $A\preceq B \implies \mathrm{Tr}A^n\le\mathrm{Tr}B^n$. We continue and write

$$\begin{aligned}
\mathrm{Tr}\widetilde{X}_{k+1}^{1-1/q} \le \mathrm{Tr}\Big(X_k^{1/q}-\frac{X_k^{1/q}PP^TX_k^{1/q}}{1+\beta}\Big)^{q-1} &= \mathrm{Tr}\Big(X_k^{1/2q}\big(I-\frac{X_k^{1/2q}PP^TX_k^{1/2q}}{1+\beta}\big)X_k^{1/2q}\Big)^{q-1}\\
&\le \mathrm{Tr}\Big(X_k^{(q-1)/2q}\big(I-\frac{X_k^{1/2q}PP^TX_k^{1/2q}}{1+\beta}\big)^{q-1}X_k^{(q-1)/2q}\Big)\\
&= \mathrm{Tr}\Big(X_k^{1-1/q}\big(I-\frac{X_k^{1/2q}PP^TX_k^{1/2q}}{1+\beta}\big)^{q-1}\Big) \enspace,
\end{aligned}$$

where the inequality again uses the Lieb-Thirring trace inequality. Denoting by $D\stackrel{\mathrm{def}}{=}\frac{X_k^{1/2q}PP^TX_k^{1/2q}}{1+\beta}\preceq\frac{\beta}{1+\beta}I$ (which uses (E.6) again), we see that

$$\big(I-D\big)^{q-1}\preceq I-(q-1)D+O(q^2\beta)\cdot D \enspace.$$

This matrix inequality can be proved by first turning into its eigenbasis, and then verifying that $(1-x)^{q-1}\le 1-(q-1)x+O(q^2\beta)x$ for all $x\in[0,\frac{\beta}{1+\beta}]$ (which uses the fact that $\beta\le 1/2q$). This concludes that

$$\begin{aligned}
\mathrm{Tr}\widetilde{X}_{k+1}^{1-1/q} &\le \mathrm{Tr}\Big(X_k^{1-1/q}\big(I-\frac{X_k^{1/2q}PP^TX_k^{1/2q}}{1+\beta}\big)^{q-1}\Big)\\
&\le \mathrm{Tr}\Big(X_k^{1-1/q}\Big(I-(q-1)\big(1-O(q\beta)\big)\frac{X_k^{1/2q}PP^TX_k^{1/2q}}{1+\beta}\Big)\Big)\\
&= \mathrm{Tr}X_k^{1-1/q}-(q-1)\big(1-O(q\beta)\big)X_k\bullet PP^T \enspace.
\end{aligned}$$

Therefore,

$$
\begin{aligned}
V_{\widetilde{X}_{k+1}}(X_k) &= \widetilde{X}_{k+1}^{-1/q} \bullet X_k + \frac{1}{q-1}\mathrm{Tr}\widetilde{X}_{k+1}^{1-1/q} - \frac{q}{q-1}\mathrm{Tr}X_k^{1-1/q} \\
&= (X_k^{-1/q} + PP^T) \bullet X_k + \frac{1}{q-1}\mathrm{Tr}\widetilde{X}_{k+1}^{1-1/q} - \frac{q}{q-1}\mathrm{Tr}X_k^{1-1/q} \\
&= PP^T \bullet X_k + \frac{1}{q-1}\big(\mathrm{Tr}\widetilde{X}_{k+1}^{1-1/q} - \mathrm{Tr}X_k^{1-1/q}\big) \\
&= O(q\beta) \cdot PP^T \bullet X_k = O(q\alpha^2)(X_k \bullet |F_k|) \cdot \|X_k^{1/2q} F_k X_k^{1/2q}\|_{\mathsf{spe}} \ .
\end{aligned}
$$

Finally, substituting the above upper bound on $V_{\widetilde{X}_{k+1}}(X_k)$ into (E.5) and telescoping it for $k = 0, \dots, T-1$, we obtain

$$
\sum_{k=0}^{T-1} \langle F_k, X_k - U \rangle \le \frac{V_{\widetilde{X}_0}(U) - V_{\widetilde{X}_T}(U)}{\alpha} + O(q\alpha) \sum_{k=0}^{T-1} (X_k \bullet |F_k|) \cdot \|X_k^{1/2q} F_k X_k^{1/2q}\|_{\mathsf{spe}} \ .
$$

The desired result of this theorem now follows from the above inequality and the simple upper bound $V_{\widetilde{X}_0}(U) = V_{X_0}(U) \le \frac{q}{q-1} n^{1/q}$ and the nonnegativity $V_{\widetilde{X}_T}(U) \ge 0$. □

# F Robust Linear-Sized Sparsification

In this section, we deduce the more generalized version of the same result presented in Section 5, with the following major differences.

- **Regularizer.** In this section, we allow the general $\ell_{1-1/q}$ regularizer to be used, for any even integer $q \ge 2$, rather than just the $\ell_{1/2}$ regularizer. (The assumption on $q$ being even integer rather than all reals no less than 2 is only for the sake of proof convenience.)
- **High rank.** In this section, we allow $\widehat{L}_e$ to be possibly of high rank, rather than just rank 1.
- **Approximate computations.** In this section, we allow many computations to be approximate rather than exact. This will enable the algorithm to be more efficiently implemented in the next section (Appendix G). In particular, we allow the following quantities to be approximately computed.
  - We only need $\mathrm{Tr}\widehat{L}_e$ to be in $[1-\varepsilon_1, 1]$ rather than exactly one.
  - We only need $\mathrm{Tr}X_k$ and $\mathrm{Tr}Y_k$ to be in $[1, 1+\varepsilon_1]$ rather than exactly one.
  - We only need $\widehat{L}_e \bullet X_k$ and $\widehat{L}_e \bullet Y_k$ to be computed only up to a $(1+\varepsilon_1)$ multiplicative error.

  We will assume throughout this paper that $\varepsilon_1 < 1/2$.

## F.1 The Problem

Suppose we are given a decomposition of the identity matrix $I = \sum_{e=1}^m w_e \widehat{L}_e$, where each $\widehat{L}_e$ satisfies ① $0 \preceq \widehat{L}_e \preceq I$, ② $\mathrm{Tr}\widehat{L}_e \in [1-\varepsilon_1, 1]$, and ③ $\widehat{L}_e$ may be of high rank. The weights $w_e > 0$ may be unknown.

In this section, we are interested in using the $\ell_{1-1/q}$ regularizer for $\texttt{MirrorDescent}_{\ell_{1-1/q}}$ in order to find scalars $s_e \ge 0$ satisfying

$$
I \preceq \sum_{e=1}^m s_e \cdot \widehat{L}_e \preceq \Big(1 + \sqrt{\frac{8q^2}{q-1}} \cdot \varepsilon + O(\varepsilon_1 + q\varepsilon^2 + \varepsilon_1 \varepsilon \sqrt{q})\Big) I \ , \qquad \text{(F.1)}
$$

while the sparsity of $s$ —that is, $|\{e \in [m] : s_e > 0\}|$— is at most $n/\varepsilon^2$. We will not worry about the running time in this section, and defer all the implementation details to Appendix G.

Throughout this section, we pick $w(X)$ to be the $\ell_{1-1/q}$ regularizer and $V_X(Y)$ to be its induced Bregman divergence.

## F.2 Our Algorithm

Maintain two sequences $X_k, Y_k \succeq 0$ satisfying $\mathrm{Tr}X_k, \mathrm{Tr}Y_k \in [1, 1+\varepsilon_1]$. At the very beginning we choose $X_0 = \frac{1}{n}I$ and $Y_0 = \frac{1}{n}I$ as before.

At each iteration $k = 0, 1, \dots, T-1$, find an arbitrary $e_k$ such that

$$\mathsf{Dot}(\widehat{L}_{e_k}, X_k) \le (1+\varepsilon_1)^2 \mathsf{Dot}(\widehat{L}_{e_k}, Y_k) \enspace ,$$

where $\mathsf{Dot}(\widehat{L}_e, X)$ is some algorithm[10] that approximately computes $\widehat{L}_e \bullet X$ and satisfies

$$\widehat{L}_e \bullet X \le \mathsf{Dot}(\widehat{L}_e, X) \le (1+\varepsilon_1) \cdot \widehat{L}_e \bullet X \enspace .$$

We can always do so because after averaging,

$$\begin{aligned}\sum_e w_e \mathsf{Dot}(\widehat{L}_e, X_k) &\le (1+\varepsilon_1) \sum_e (w_e \widehat{L}_e) \bullet X_k = (1+\varepsilon_1)\mathrm{Tr}X_k \\ &\le (1+\varepsilon_1)^2 \mathrm{Tr}Y_k = (1+\varepsilon_1)^2 \sum_e (w_e \widehat{L}_e) \bullet Y_k \le (1+\varepsilon_1)^2 \sum_e w_e \mathsf{Dot}(\widehat{L}_e, Y_k) \enspace .\end{aligned}$$

At each iteration $k = 0, 1, \dots, T-1$, we perform updates by finding[11] arbitrary $\delta_X, \delta_Y \ge 0$ satisfying

$$Y_k^{-1/q} + \frac{\alpha \widehat{L}_{e_k}}{\mathsf{Dot}(\widehat{L}_{e_k}, Y_k)^{1/q}} - \delta_Y I \succeq 0 \quad \text{and} \quad \mathrm{Tr}X_{k+1}, \mathrm{Tr}Y_{k+1} \in [1, 1+\varepsilon_1] \enspace ,$$

where

$$X_{k+1} \stackrel{\text{def}}{=} \Big( X_k^{-1/q} + \frac{-\alpha \widehat{L}_{e_k}}{\mathsf{Dot}(\widehat{L}_{e_k}, X_k)^{1/q}} + \delta_X I \Big)^{-q} \quad \text{and} \quad Y_{k+1} \stackrel{\text{def}}{=} \Big( Y_k^{-1/q} + \frac{\alpha \widehat{L}_{e_k}}{\mathsf{Dot}(\widehat{L}_{e_k}, Y_k)^{1/q}} - \delta_Y I \Big)^{-q} \enspace .$$

Above, $\alpha > 0$ is some parameter that will be specified at the end of this section. Note that this corresponds to performing updates

$$\text{“} \quad X_{k+1} \leftarrow \operatorname*{arg\,min}_{Z \in \Delta_{n\times n}} \Big\{ V_{X_k}(Z) + \Big\langle \frac{-\alpha \widehat{L}_{e_k}}{\mathsf{Dot}(\widehat{L}_{e_k}, X_k)^{1/q}}, Z \Big\rangle \Big\} \quad \text{”} \qquad \text{and}$$

$$\text{“} \quad Y_{k+1} \leftarrow \operatorname*{arg\,min}_{Z \in \Delta_{n\times n}} \Big\{ V_{Y_k}(Z) + \Big\langle \frac{\alpha \widehat{L}_{e_k}}{\mathsf{Dot}(\widehat{L}_{e_k}, Y_k)^{1/q}}, Z \Big\rangle \Big\} \quad \text{”}$$

however, we have **not** required $\mathrm{Tr}X_{k+1} = \mathrm{Tr}Y_{k+1}$ to be precisely equal to 1.

For analysis purpose only, we also define $\widetilde{X}_{k+1}$ and $\widetilde{Y}_{k+1}$ to be similar updates but without $\delta_X$ or $\delta_Y$:

$$\widetilde{X}_{k+1} \stackrel{\text{def}}{=} \Big( X_k^{-1/q} + \frac{-\alpha \widehat{L}_{e_k}}{\mathsf{Dot}(\widehat{L}_{e_k}, X_k)^{1/q}} \Big)^{-q} \quad \text{and} \quad \widetilde{Y}_{k+1} \stackrel{\text{def}}{=} \Big( Y_k^{-1/q} + \frac{\alpha \widehat{L}_{e_k}}{\mathsf{Dot}(\widehat{L}_{e_k}, Y_k)^{1/q}} \Big)^{-q} \enspace .$$

We assume also $\widetilde{X}_0 \stackrel{\text{def}}{=} X_0$.

Note that $\widetilde{Y}_{k+1}$ is always well defined. Claim F.1 below shows that as long as $\alpha < 1$, it always satisfies $X_k^{-1/q} + \frac{-\alpha \widehat{L}_{e_k}}{\mathsf{Dot}(\widehat{L}_{e_k}, X_k)^{1/q}} \succeq 0$, so $\widetilde{X}_{k+1}$ is also well defined.

**Claim F.1.** *For every $e \in [m]$, we have $X_k^{-1/q} \succeq \frac{\widehat{L}_e}{(\widehat{L}_e \bullet X_k)^{1/q}} \succeq \frac{\widehat{L}_e}{\mathsf{Dot}(\widehat{L}_e, X_k)^{1/q}}$. In addition, denoting by $\frac{\alpha \widehat{L}_e}{\mathsf{Dot}(\widehat{L}_e, X_k)^{1/q}} = PP^T$, we have $0 \preceq P^T X_k^{1/q} P \preceq \alpha I$.*

---

[10] The implementation of this algorithm will be described in Appendix G.

[11] The existence of such $\delta_X$ and $\delta_Y$ shall become soon (due to Claim F.1). The implementation of these updates will be described in Appendix G.

*Similarly, for every $e \in [m]$, we have $Y_k^{-1/q} \succeq \frac{\widehat{L}_e}{(\widehat{L}_e \bullet Y_k)^{1/q}} \succeq \frac{\widehat{L}_e}{\mathsf{Dot}(\widehat{L}_e, Y_k)^{1/q}}$. In addition, denoting by $\frac{\alpha \widehat{L}_e}{\mathsf{Dot}(\widehat{L}_e, Y_k)^{1/q}} = PP^T$, we have $0 \preceq P^T Y_k^{1/q} P \preceq \alpha I$.*

*Proof.* We only prove the $X_k$ part because the $Y_k$ part is similar. We first compute

$$\|X_k^{1/2q} \widehat{L}_e X_k^{1/2q}\|_{\mathsf{spe}}^q \leq \mathrm{Tr}((X_k^{1/2q} \widehat{L}_e X_k^{1/2q})^q) \leq \mathrm{Tr}(X_k^{1/2} (\widehat{L}_e)^q X_k^{1/2}) \ ,$$

where the last inequality follows from the Lieb-Thirring trace inequality.

Next, using the fact that $\widehat{L}_e \preceq I$, we obtain that $(\widehat{L}_e)^q \preceq \widehat{L}_e$. Therefore,

$$\|X_k^{1/2q} \widehat{L}_e X_k^{1/2q}\|_{\mathsf{spe}}^q \leq \mathrm{Tr}(X_k^{1/2} \widehat{L}_e X_k^{1/2}) = \widehat{L}_e \bullet X_k \ .$$

In other words, we have $X_k^{1/2q} \widehat{L}_e X_k^{1/2q} \preceq (\widehat{L}_e \bullet X_k)^{1/q} \cdot I$ which means $X_k^{-1/q} \succeq \frac{\widehat{L}_e}{(\widehat{L}_e \bullet X_k)^{1/q}}$. We automatically have $\frac{\widehat{L}_e}{(\widehat{L}_e \bullet X_k)^{1/q}} \succeq \frac{\widehat{L}_e}{\mathsf{Dot}(\widehat{L}_e, X_k)^{1/q}}$ because $\mathsf{Dot}(\widehat{L}_e, X_k) \geq \widehat{L}_e \bullet X_k$.

To prove the second half, beginning from $X_k^{-1/q} \succeq \frac{1}{\alpha} \cdot PP^T$, we left multiply it with $P^T X_k^{1/q}$ and right multiply it with $X_k^{1/q} P$, and obtain $P^T X_k^{1/q} P \succeq \frac{1}{\alpha} \cdot P^T X_k^{1/q} P P^T X_k^{1/q} P$. Denoting by $D \stackrel{\text{def}}{=} P^T X_k^{1/q} P$, we have $D \succeq \frac{1}{\alpha} D^2$, which immediately implies $0 \preceq D \preceq \alpha I$ as desired. □

We have now finished the description of the algorithm. We remark here that $\mathrm{Tr}\widetilde{X}_{k+1} < \mathrm{Tr}X_k$ and $\mathrm{Tr}\widetilde{Y}_{k+1} > \mathrm{Tr}Y_k$. Therefore, since $\mathrm{Tr}X_{k+1}$ increases as $\delta_X$ increases, while $\mathrm{Tr}Y_{k+1}$ decreases as $\delta_Y$ increase, we conclude the existence of $\delta_X, \delta_Y \geq 0$ so that $\mathrm{Tr}X_{k+1}, \mathrm{Tr}Y_{k+1} \in [1, 1+\varepsilon_1]$.

## F.3 Our Analysis

We begin by reproving essentially the first half of Theorem 3.2: that is, to prove (E.3). We need to pay extra attention here since our $\mathrm{Tr}X_k$ and $\mathrm{Tr}Y_k$ do not precisely equal to 1.

**Lemma F.2.** *For every $U_X \succeq 0$ satisfying $\mathrm{Tr}U_X \leq 1$, and every $U_Y \succeq 0$ satisfying $\mathrm{Tr}U_Y \geq 1 + \varepsilon_1$,*

$$\Big\langle \frac{-\alpha \widehat{L}_{e_k}}{\mathsf{Dot}(\widehat{L}_{e_k}, X_k)^{1/q}}, X_k - U_X \Big\rangle \leq V_{\widetilde{X}_{k+1}}(X_k) + V_{\widetilde{X}_k}(U_X) - V_{\widetilde{X}_{k+1}}(U_X) \ , \quad \textit{and}$$

$$\Big\langle \frac{\alpha \widehat{L}_{e_k}}{\mathsf{Dot}(\widehat{L}_{e_k}, Y_k)^{1/q}}, Y_k - U_Y \Big\rangle \leq V_{\widetilde{Y}_{k+1}}(Y_k) + V_{\widetilde{Y}_k}(U_Y) - V_{\widetilde{Y}_{k+1}}(U_Y) \ .$$

*Proof.* We first prove the $X_k$ part. By our choice of the regularizer, we have

$$0 = \nabla w(\widetilde{X}_{k+1}) - \nabla w(X_k) + \frac{-\alpha \widehat{L}_{e_k}}{\mathsf{Dot}(\widehat{L}_{e_k}, X_k)^{1/q}} = -\widetilde{X}_{k+1}^{-1/q} + X_k^{-1/q} + \frac{-\alpha \widehat{L}_{e_k}}{\mathsf{Dot}(\widehat{L}_{e_k}, X_k)^{1/q}} \ .$$

Next, we obtain that

$$\begin{aligned}
\langle \frac{-\alpha \widehat{L}_{e_k}}{\mathsf{Dot}(\widehat{L}_{e_k}, X_k)^{1/q}}, X_k - U_X \rangle &= \langle \nabla w(X_k) - \nabla w(\widetilde{X}_{k+1}), X_k - U_X \rangle \\
&\stackrel{①}{=} V_{X_k}(U_X) - V_{\widetilde{X}_{k+1}}(U_X) + V_{\widetilde{X}_{k+1}}(X_k) \\
&\stackrel{②}{\leq} V_{\widetilde{X}_k}(U_X) - V_{\widetilde{X}_{k+1}}(U_X) + V_{\widetilde{X}_{k+1}}(X_k) \ .
\end{aligned}$$

Above, ① is due to the three-point equality of Bregman divergence, and ② comes from

$$V_{X_k}(U_X) - V_{\widetilde{X}_k}(U_X) \overset{③}{=} \big(X_k^{-1/q} - \widetilde{X}_k^{-1/q}\big) \bullet U_X + \frac{1}{q-1}\big(\mathrm{Tr}X_k^{1-1/q} - \mathrm{Tr}\widetilde{X}_k^{1-1/q}\big)$$
$$\overset{④}{=} \delta_X \mathrm{Tr}U_X + \frac{1}{q-1}\sum_i \frac{1}{\lambda_i^{q-1}} - \frac{1}{(\lambda_i - \delta_X)^{q-1}}$$
$$\overset{⑤}{\leq} \delta_X \mathrm{Tr}U_X - \delta_X \sum_i \frac{1}{\lambda_i^q} \overset{⑥}{\leq} 0 \enspace.$$

Here, ③ is owing to the definition of Bregman divergence. ④ comes from the fact that $\widetilde{X}_{k+1}^{-1/q} = X_{k+1}^{-1/q} - \delta_X I$, and the definition of choosing $\lambda_i$ to be the $i$-th eigenvalue of $X_{k+1}^{-1/q}$. ⑤ follows from the convexity of $f(x) = x^{1-q}$ which implies $f(\lambda_i) - f(\lambda_i - \delta_X) \leq \nabla f(\lambda_i)\cdot\delta_X$. ⑥ is by our assumption of $\mathrm{Tr}U_X \leq 1$ as well as $\mathrm{Tr}X_{k+1} = \sum_i \frac{1}{\lambda_i^q} \geq 1$.

Similarly, for the $Y_k$ part, we can compute

$$\langle \frac{\alpha \widehat{L}_{e_k}}{\mathsf{Dot}(\widehat{L}_{e_k}, Y_k)^{1/q}}, Y_k - U_Y\rangle = \langle \nabla w(Y_k) - \nabla w(\widetilde{Y}_{k+1}), Y_k - U_Y\rangle$$
$$\overset{①}{=} V_{Y_k}(U_Y) - V_{\widetilde{Y}_{k+1}}(U_Y) + V_{\widetilde{Y}_{k+1}}(Y_k)$$
$$\overset{②}{\leq} V_{\widetilde{Y}_k}(U_Y) - V_{\widetilde{Y}_{k+1}}(U_Y) + V_{\widetilde{Y}_{k+1}}(Y_k) \enspace.$$

Above, ① is due to the three-point equality, and inequality ② comes from

$$V_{Y_k}(U_Y) - V_{\widetilde{Y}_k}(U_Y) \overset{③}{=} \big(Y_k^{-1/q} - \widetilde{Y}_k^{-1/q}\big) \bullet U_Y + \frac{1}{q-1}\big(\mathrm{Tr}Y_k^{1-1/q} - \mathrm{Tr}\widetilde{Y}_k^{1-1/q}\big)$$
$$\overset{④}{=} -\delta_Y \mathrm{Tr}U_Y + \frac{1}{q-1}\sum_i \frac{1}{\lambda_i^{q-1}} - \frac{1}{(\lambda_i + \delta_Y)^{q-1}}$$
$$\overset{⑤}{\leq} -\delta_Y \mathrm{Tr}U_Y + \delta_Y \sum_i \frac{1}{\lambda_i^q} \overset{⑥}{\leq} 0 \enspace.$$

Here, ③ is owing to the definition of Bregman divergence. ④ comes from the fact that $\widetilde{Y}_{k+1}^{-1/q} = Y_{k+1}^{-1/q} + \delta_Y I$, and the definition of choosing $\lambda_i$ to be the $i$-th eigenvalue of $Y_{k+1}^{-1/q}$. ⑤ follows from the convexity of $f(x) = x^{1-q}$ which implies $f(\lambda_i) - f(\lambda_i + \delta_Y) \leq \nabla f(\lambda_i)\cdot(-\delta_Y)$. ⑥ is by our assumption of $\mathrm{Tr}U_Y \geq 1 + \varepsilon_1$ as well as $\mathrm{Tr}Y_{k+1} = \sum_i \frac{1}{\lambda_i^q} \leq 1 + \varepsilon_1$. □

In a next step, we reprove essentially the second half of Theorem 3.2: that is, to provide upper bounds on $V_{\widetilde{X}_{k+1}}(X_k)$ and $V_{\widetilde{Y}_{k+1}}(Y_k)$ in Lemma F.3 and Lemma F.4.

**Lemma F.3.** *As long as $q \geq 2$ and $\alpha \leq 1/2q$, we have*

$$V_{\widetilde{X}_{k+1}}(X_k) \leq \frac{q}{2}(\alpha^2 + O(q\alpha^3)) \cdot \big(\widehat{L}_{e_k} \bullet X_k\big)^{1-1/q} \enspace.$$

*Proof.* Suppose $\frac{\alpha \widehat{L}_{e_k}}{\mathsf{Dot}(\widehat{L}_{e_k}, X_k)^{1/q}} = PP^T$. Then, using the Sherman-Morrison-Woodbury formula,

$$\mathrm{Tr}\widetilde{X}_{k+1}^{1-1/q} = \mathrm{Tr}\big((X_k^{-1/q} - PP^T)^{-1}\big)^{q-1} = \mathrm{Tr}\Big(X_k^{1/q} + X_k^{1/q}P(I - P^T X_k^{1/q} P)^{-1} P^T X_k^{1/q}\Big)^{q-1}$$
$$\leq \mathrm{Tr}\Big(X_k^{1/q} + \frac{X_k^{1/q} PP^T X_k^{1/q}}{1-\alpha}\Big)^{q-1} \enspace,$$

where the last inequality follows because $(I - P^T X_k^{1/q} P)^{-1} \preceq \frac{1}{1-\alpha} I$ owing to Claim F.1, as well as $A \preceq B \implies \mathrm{Tr}A^n \le \mathrm{Tr}B^n$. We continue and write

$$\begin{aligned}
\mathrm{Tr}\widetilde{X}_{k+1}^{1-1/q} \le \mathrm{Tr}\Big(X_k^{1/q} + \frac{X_k^{1/q} PP^T X_k^{1/q}}{1-\alpha}\Big)^{q-1} &= \mathrm{Tr}\Big(X_k^{1/2q}\big(I + \frac{X_k^{1/2q} PP^T X_k^{1/2q}}{1-\alpha}\big)X_k^{1/2q}\Big)^{q-1} \\
&\le \mathrm{Tr}\Big(X_k^{(q-1)/2q}\big(I + \frac{X_k^{1/2q} PP^T X_k^{1/2q}}{1-\alpha}\big)^{q-1} X_k^{(q-1)/2q}\Big) \\
&= \mathrm{Tr}\Big(X_k^{1-1/q}\big(I + \frac{X_k^{1/2q} PP^T X_k^{1/2q}}{1-\alpha}\big)^{q-1}\Big) \ ,
\end{aligned}$$

where the inequality uses the Lieb-Thirring trace inequality (which relies on the fact that $q-1 \ge 1$). Finally, denoting by $D \stackrel{\text{def}}{=} \frac{X_k^{1/2q} PP^T X_k^{1/2q}}{1-\alpha} \preceq \frac{\alpha}{1-\alpha} I$, we see that

$$\big(I + D\big)^{q-1} \preceq I + (q-1)D + \Big(\frac{(q-1)(q-2)}{2}\alpha + O(q^3\alpha^2)\Big)D \ .$$

This above matrix inequality can be proved by first turning into its eigenbasis, and then verifying that $(1+x)^{q-1} \le 1 + (q-1)x + \frac{(q-1)(q-2)}{2}\alpha x + O(q^3\alpha^2)x$ for all $x \in [0, \frac{\alpha}{1-\alpha}]$. (This uses the fact that $\alpha \le 1/2q$). Next, using the above matrix inequality, we conclude that

$$\begin{aligned}
\mathrm{Tr}\widetilde{X}_{k+1}^{1-1/q} &\le \mathrm{Tr}\Big(X_k^{1-1/q}\big(I + \frac{X_k^{1/2q} PP^T X_k^{1/2q}}{1-\alpha}\big)^{q-1}\Big) \\
&\le \mathrm{Tr}\Big(X_k^{1-1/q}\Big(I + \Big((q-1) + \frac{(q-1)(q-2)}{2}\alpha + O(q^3\alpha^2)\Big)\frac{X_k^{1/2q} PP^T X_k^{1/2q}}{1-\alpha}\Big)\Big) \\
&= \mathrm{Tr}X_k^{1-1/q} + (q-1)\frac{1 + \frac{q-2}{2}\alpha + O(q^2\alpha^2)}{1-\alpha} X_k \bullet PP^T \ .
\end{aligned}$$

Therefore,

$$\begin{aligned}
V_{\widetilde{X}_{k+1}}(X_k) &= \widetilde{X}_{k+1}^{-1/q} \bullet X_k + \frac{1}{q-1}\mathrm{Tr}\widetilde{X}_{k+1}^{1-1/q} - \frac{q}{q-1}\mathrm{Tr}X_k^{1-1/q} \\
&= (X_k^{-1/q} - PP^T) \bullet X_k + \frac{1}{q-1}\mathrm{Tr}\widetilde{X}_{k+1}^{1-1/q} - \frac{q}{q-1}\mathrm{Tr}X_k^{1-1/q} \\
&= -PP^T \bullet X_k + \frac{1}{q-1}\big(\mathrm{Tr}\widetilde{X}_{k+1}^{1-1/q} - \mathrm{Tr}X_k^{1-1/q}\big) \\
&\le PP^T \bullet X_k\Big(-1 + \frac{1 + \frac{q-2}{2}\alpha + O(q^2\alpha^2)}{1-\alpha}\Big) \\
&= \frac{q}{2}(\alpha + O(q\alpha^2)) \cdot PP^T \bullet X_k \\
&\le \frac{q}{2}(\alpha^2 + O(q\alpha^3)) \cdot \big(\widehat{L}_{e_k} \bullet X_k\big)^{1-1/q} \ .
\end{aligned}$$

□

**Lemma F.4.** *As long as $q \ge 2$ and $\alpha \le 1/2q$, we have*

$$V_{\widetilde{Y}_{k+1}}(Y_k) \le \frac{q}{2}\big(\alpha^2 + O(\alpha^3)\big) \cdot \big(\widehat{L}_{e_k} \bullet Y_k\big)^{1-1/q} \ .$$

*Proof.* Suppose $\frac{\alpha \widehat{L}_{e_k}}{\mathsf{Dot}(\widehat{L}_{e_k}, Y_k)^{1/q}} = PP^T$. Then, using the Sherman-Morrison-Woodbury formula,

$$\begin{aligned}
\mathrm{Tr}\widetilde{Y}_{k+1}^{1-1/q} = \mathrm{Tr}\big((Y_k^{-1/q} + PP^T)^{-1}\big)^{q-1} &= \mathrm{Tr}\Big(Y_k^{1/q} - Y_k^{1/q}P(I + P^T Y_k^{1/q} P)^{-1} P^T Y_k^{1/q}\Big)^{q-1} \\
&\le \mathrm{Tr}\Big(Y_k^{1/q} - \frac{Y_k^{1/q} PP^T Y_k^{1/q}}{1+\alpha}\Big)^{q-1} \ ,
\end{aligned}$$

where the last inequality follows because $(I + P^T Y_k^{1/q} P)^{-1} \succeq \frac{1}{1+\alpha} I$ owing to Claim F.1, as well as $A \preceq B \implies \mathrm{Tr} A^n \leq \mathrm{Tr} B^n$. We continue and write

$$\begin{aligned}
\mathrm{Tr} \widetilde{Y}_{k+1}^{1-1/q} \leq \mathrm{Tr}\Big( Y_k^{1/q} - \frac{Y_k^{1/q} P P^T Y_k^{1/q}}{1+\alpha} \Big)^{q-1} &= \mathrm{Tr}\Big( Y_k^{1/2q} \big( I - \frac{Y_k^{1/2q} P P^T Y_k^{1/2q}}{1+\alpha} \big) Y_k^{1/2q} \Big)^{q-1} \\
&\leq \mathrm{Tr}\Big( Y_k^{(q-1)/2q} \big( I - \frac{Y_k^{1/2q} P P^T Y_k^{1/2q}}{1+\alpha} \big)^{q-1} Y_k^{(q-1)/2q} \Big) \\
&= \mathrm{Tr}\Big( Y_k^{1-1/q} \big( I - \frac{Y_k^{1/2q} P P^T Y_k^{1/2q}}{1+\alpha} \big)^{q-1} \Big) \enspace,
\end{aligned}$$

where the inequality again uses the Lieb-Thirring trace inequality (which relies on the fact that $q-1 \geq 1$). Denoting by $D \stackrel{\mathrm{def}}{=} \frac{Y_k^{1/2q} P P^T Y_k^{1/2q}}{1+\alpha} \preceq \frac{\alpha}{1+\alpha} I$, we see that

$$\big( I - D \big)^{q-1} \preceq I - (q-1) D + \frac{(q-1)(q-2)\alpha}{2(1+\alpha)} D \enspace.$$

This above matrix inequality can be proved by first turning into its eigenbasis, and then verifying that $(1-x)^{q-1} \leq 1 - (q-1)x + \frac{(q-1)(q-2)}{2} \frac{\alpha}{1+\alpha} x$ for all $x \in [0, \frac{\alpha}{1+\alpha}]$. (This uses the fact that $\alpha \leq 1/2q$). Next, using the above matrix inequality, we conclude that

$$\begin{aligned}
\mathrm{Tr} \widetilde{Y}_{k+1}^{1-1/q} &\leq \mathrm{Tr}\Big( Y_k^{1-1/q} \big( I - \frac{Y_k^{1/2q} P P^T Y_k^{1/2q}}{1+\alpha} \big)^{q-1} \Big) \\
&\leq \mathrm{Tr}\Big( Y_k^{1-1/q} \Big( I - (q-1) \big( 1 - \frac{(q-2)\alpha}{2(1+\alpha)} \big) \frac{Y_k^{1/2q} P P^T Y_k^{1/2q}}{1+\alpha} \Big) \Big) \\
&= \mathrm{Tr} Y_k^{1-1/q} - (q-1) \big( 1 - \frac{(q-2)\alpha}{2(1+\alpha)} \big) \frac{1}{1+\alpha} Y_k \bullet P P^T \enspace.
\end{aligned}$$

Therefore,

$$\begin{aligned}
V_{\widetilde{Y}_{k+1}}(Y_k) &= \widetilde{Y}_{k+1}^{-1/q} \bullet Y_k + \frac{1}{q-1} \mathrm{Tr} \widetilde{Y}_{k+1}^{1-1/q} - \frac{q}{q-1} \mathrm{Tr} Y_k^{1-1/q} \\
&= (Y_k^{-1/q} + P P^T) \bullet Y_k + \frac{1}{q-1} \mathrm{Tr} \widetilde{Y}_{k+1}^{1-1/q} - \frac{q}{q-1} \mathrm{Tr} Y_k^{1-1/q} \\
&= P P^T \bullet Y_k + \frac{1}{q-1} \big( \mathrm{Tr} \widetilde{Y}_{k+1}^{1-1/q} - \mathrm{Tr} Y_k^{1-1/q} \big) \\
&\leq P P^T \bullet Y_k \Big( 1 - \frac{\big( 1 - \frac{(q-2)\alpha}{2(1+\alpha)} \big)}{1+\alpha} \Big) \\
&= \frac{q}{2} \big( \alpha + O(\alpha^2) \big) \cdot P P^T \bullet Y_k \\
&\leq \frac{q}{2} \big( \alpha^2 + O(\alpha^3) \big) \cdot \big( \widehat{L}_{e_k} \bullet Y_k \big)^{1-1/q} \enspace. \qquad \square
\end{aligned}$$

**Theorem F.5.** *Suppose $\varepsilon < \frac{1}{4\sqrt{q}}$ and $\varepsilon_1 < \frac{1}{2}$, and we choose $\alpha = \frac{\varepsilon\sqrt{2}}{\sqrt{q-1}}$ and $T = \frac{n}{\varepsilon^2}$. Then, the matrix $M_Y \stackrel{\mathrm{def}}{=} \sum_{k=0}^{T-1} \frac{\widehat{L}_{e_k}}{\mathsf{Dot}(\widehat{L}_{e_k}, Y_k)^{1/q}}$ satisfies that*

$$\lambda_{\mathsf{max}}(M_Y) - \lambda_{\mathsf{min}}(M_Y) \leq \lambda_{\mathsf{min}}(M_Y) \cdot \Big( \sqrt{\frac{8q^2}{q-1}} \cdot \varepsilon + O(\varepsilon_1 + q\varepsilon^2) \Big) \enspace.$$

This theorem provides the sparsification guarantee to our Theorem 1 and 2. We shall provide its running time guarantee in the next section.

*Proof.* Define matrices $M_X \stackrel{\text{def}}{=} \sum_{k=0}^{T-1} \frac{\widehat{L}_{e_k}}{\mathsf{Dot}(\widehat{L}_{e_k}, X_k)^{1/q}}$ and $M_Y \stackrel{\text{def}}{=} \sum_{k=0}^{T-1} \frac{\widehat{L}_{e_k}}{\mathsf{Dot}(\widehat{L}_{e_k}, Y_k)^{1/q}}$. Also, denote by $\xi \stackrel{\text{def}}{=} \frac{q}{2}(\alpha + O(q\alpha^2))$.

We are now ready to rededuce (5.3) and (5.4) in Section 5.

Combining Lemma F.2 and Lemma F.3, and telescoping for $k = 0, 1, \ldots, T-1$, we have

$$\forall U_X \succeq 0 \text{ satisfying } \mathrm{Tr} U_X = 1, \quad M_X \bullet U_X \leq \frac{V_{\widetilde{X}_0}(U_X)}{\alpha} + (1+\xi) \sum_{k=0}^{T-1} \big(\widehat{L}_{e_k} \bullet X_k\big)^{1-1/q} \tag{F.2}$$

$$\leq \frac{q n^{1/q}}{(q-1)\alpha} + (1+\xi) \sum_{k=0}^{T-1} (\widehat{L}_{e_k} \bullet X_k)^{1-1/q} \ . \tag{F.3}$$

Above, the second inequality uses the fact that $V_{\widetilde{X}_0}(U_X) \leq \frac{q}{q-1} n^{1/q}$.

Combining Lemma F.2 and Lemma F.4, and telescoping for $k = 0, 1, \ldots, T-1$, we have

$$\forall U_Y \succeq 0, \mathrm{Tr} U_Y = 1 + \varepsilon_1, \quad M_Y \bullet U_Y \geq -\frac{V_{\widetilde{Y}_0}(U_Y)}{\alpha} + (1-\xi) \sum_{k=0}^{T-1} (\widehat{L}_{e_k} \bullet Y_k)^{1-1/q}$$

$$\geq -\frac{q(1+\varepsilon_1) n^{1/q}}{(q-1)\alpha} + (1-\xi) \sum_{k=0}^{T-1} \big(\widehat{L}_{e_k} \bullet Y_k\big)^{1-1/q} \ . \tag{F.4}$$

Above, the second inequality uses the fact that $V_{\widetilde{Y}_0}(U_Y) \leq \frac{q(1+\varepsilon_1)}{q-1} n^{1/q}$.

Similar to the proof in Section 5, we provide deduce our eigenvalue inequality in two steps.

**Lowerbounding $\boldsymbol{\lambda_{\min}(M_Y)}$.** Since we have assumed each $\mathrm{Tr}\widehat{L}_e$ to be at least $1 - \varepsilon_1$, we have

$$\mathrm{Tr}(M_X) = \sum_{k=0}^{T-1} \frac{\mathrm{Tr}\widehat{L}_{e_k}}{\mathsf{Dot}(\widehat{L}_{e_k}, X_k)^{1/q}} \geq \frac{1-\varepsilon_1}{(1+\varepsilon_1)^{1/q}} \sum_{k=0}^{T-1} \frac{1}{(\widehat{L}_{e_k} \bullet X_k)^{1/q}} \ .$$

Denoting by $a_k = \widehat{L}_{e_k} \bullet X_k$, we can write $\mathrm{Tr}(M_X) \geq \frac{1-\varepsilon_1}{(1+\varepsilon_1)^{1/q}} \sum_{k=0}^{T-1} \frac{1}{a_k^{1/q}}$. Applying (F.2) with the choice of $U_X = \frac{1}{n} I = X_0$, we have

$$\frac{1-\varepsilon_1}{n(1+\varepsilon_1)^{1/q}} \sum_{k=0}^{T-1} \frac{1}{a_k^{1/q}} \leq \frac{1}{n} \mathrm{Tr} M_X = M_X \bullet U_X \leq (1+\xi) \sum_{k=0}^{T-1} (\widehat{L}_{e_k} \bullet X_k)^{1-1/q} \leq (1+\xi) \sum_{k=0}^{T-1} a_k^{1-1/q} \ .$$

Using the above inequality we obtain

$$\sum_{k=0}^{T-1} a_k^{1-1/q} \geq \frac{1}{\big(n(1+\xi)(1+\varepsilon_1)^{1/q}(1-\varepsilon_1)^{-1}\big)^{1-1/q}} \Big(\sum_{k=0}^{T-1} a_k^{1-1/q}\Big)^{1/q} \Big(\sum_{k=0}^{T-1} \frac{1}{a_k^{1/q}}\Big)^{1-1/q}$$

$$\geq \frac{T}{n^{1-1/q}(1+\xi)^{1-1/q}(1+\varepsilon_1)^{1/q-1/q^2}(1-\varepsilon_1)^{1/q-1}} \ ,$$

where the last inequality follows from Hölder's inequality. If we choose $T = \frac{n}{\varepsilon^2}$, this immediately gives

$$\sum_{k=0}^{T-1} (\widehat{L}_{e_k} \bullet X_k)^{1-1/q} = \sum_{k=0}^{T-1} a_k^{1-1/q} \geq \frac{n^{1/q}}{\varepsilon^2} (1 - O(q\alpha + \varepsilon_1)) \ . \tag{F.5}$$

Finally, substituting (F.5) into (F.4), and choosing $U_Y$ so that $M_Y \bullet U_Y = (1+\varepsilon_1)\lambda_{\min}(M_Y)$,

we have

$$
\begin{aligned}
(1+\varepsilon_1)\lambda_{\mathsf{min}}(M_Y) &\geq -\frac{q(1+\varepsilon_1)n^{1/q}}{(q-1)\alpha} + (1-\xi)\frac{1}{(1+\varepsilon_1)^{3-3/q}}\sum_{k=0}^{T-1}(\widehat{L}_{e_k} \bullet X_k)^{1-1/q} \\
&\geq -\frac{2qn^{1/q}}{(q-1)\alpha} + (1-\xi)\frac{1}{(1+\varepsilon_1)^{3-3/q}}\frac{n^{1/q}}{\varepsilon^2}(1-O(q\alpha+\varepsilon_1)) \\
&\geq -\frac{2qn^{1/q}}{(q-1)\alpha} + \frac{n^{1/q}}{\varepsilon^2}(1-O(q\alpha+\varepsilon_1)) \\
&\geq \frac{n^{1/q}}{\varepsilon^2}(1-O(q\alpha+\varepsilon_1+\varepsilon^2/\alpha)) \ . \qquad \text{(F.6)}
\end{aligned}
$$

Above, the first inequality is due to our choice of $e_k$ which satisfies

$$(1+\varepsilon_1)^3\widehat{L}_{e_k} \bullet Y_k \geq (1+\varepsilon_1)^2\mathsf{Dot}(\widehat{L}_{e_k}, Y_k) \geq \mathsf{Dot}(\widehat{L}_{e_k}, X_k) \geq \widehat{L}_{e_k} \bullet X_k \ . \qquad \text{(F.7)}$$

**Upper bounding $\lambda_{\mathsf{max}}(M_Y) - \lambda_{\mathsf{min}}(M_Y)$.** This time, combining (F.3) and (F.4), as well as using (F.7), we compute that

$$\frac{1}{1+\xi}\big(M_Y \bullet U_X - \frac{qn^{1/q}}{(q-1)\alpha}\big) \leq \frac{1}{1+\xi}\big(M_X \bullet U_X - \frac{qn^{1/q}}{(q-1)\alpha}\big) \leq \frac{(1+\varepsilon_1)^{3-3/q}}{1-\xi}\big(M_Y \bullet U_Y + \frac{q(1+\varepsilon_1)n^{1/q}}{(q-1)\alpha}\big) \ .$$

Choosing $U_X$ so that $M_Y \bullet U_X = \lambda_{\mathsf{max}}(M_Y)$, and $U_Y$ so that $M_Y \bullet U_Y = (1+\varepsilon_1)\lambda_{\mathsf{min}}(M_Y)$, we can rewrite the above inequality as

$$\frac{1}{1+\xi}\big(\lambda_{\mathsf{max}}(M_Y) - \frac{qn^{1/q}}{(q-1)\alpha}\big) \leq \frac{(1+\varepsilon_1)^{3-3/q}}{1-\xi}(1+\varepsilon_1)\big(\lambda_{\mathsf{min}}(M_Y) + \frac{qn^{1/q}}{(q-1)\alpha}\big) \ . \qquad \text{(F.8)}$$

To turn this joint multiplicative-additive error into a purely multiplicative one, we further rewrite it as

$$
\begin{aligned}
\lambda_{\mathsf{max}}(M_Y) - \lambda_{\mathsf{min}}(M_Y) &\leq \frac{2\xi + O(\varepsilon_1)}{1-\xi}\lambda_{\mathsf{min}}(M_Y) + \frac{1+\xi+O(\varepsilon_1)}{1-\xi}\frac{qn^{1/q}}{(q-1)\alpha} + \frac{qn^{1/q}}{(q-1)\alpha} \\
&= \frac{2\xi + O(\varepsilon_1)}{1-\xi}\lambda_{\mathsf{min}}(M_Y) + \frac{2q}{q-1}\frac{1+O(\varepsilon_1)}{1-\xi}\frac{n^{1/q}}{\alpha} \\
&\leq \frac{2\xi + O(\varepsilon_1)}{1-\xi}\lambda_{\mathsf{min}}(M_Y) + \frac{2q}{q-1}\cdot\lambda_{\mathsf{min}}(M_Y)\frac{\varepsilon^2}{\alpha}(1+O(q\alpha+\varepsilon_1+\varepsilon^2/\alpha)) \\
&= \lambda_{\mathsf{min}}(M_Y)\cdot\Big(q\alpha + \frac{2q}{q-1}\frac{\varepsilon^2}{\alpha} + O(\varepsilon_1 + q\varepsilon^2 + \varepsilon_1\varepsilon^2/\alpha + \varepsilon^4/\alpha^2 + q^2\alpha^2)\Big) \ .
\end{aligned}
$$

Above, the second inequality uses (F.6). Now, it is clear that by choosing $\alpha = \frac{\varepsilon\sqrt{2}}{\sqrt{q-1}} \leq \frac{1}{2q}$, we have

$$
\begin{aligned}
\lambda_{\mathsf{max}}(M_Y) - \lambda_{\mathsf{min}}(M_Y) &\leq \lambda_{\mathsf{min}}(M_Y)\cdot\Big(\sqrt{\frac{8q^2}{q-1}}\cdot\varepsilon + O(\varepsilon_1 + q\varepsilon^2 + \varepsilon_1\varepsilon\sqrt{q})\Big) \\
&\leq \lambda_{\mathsf{min}}(M_Y)\cdot\Big(\sqrt{\frac{8q^2}{q-1}}\cdot\varepsilon + O(\varepsilon_1 + q\varepsilon^2)\Big) \ . \qquad \square
\end{aligned}
$$

## F.4 An Additional Property

Recall that in the previous subsection, we have constructed $M_X$ and $M_Y$ and proved that $\lambda_{\mathsf{min}}(M_Y)$ (and in fact $\lambda_{\mathsf{min}}(M_Y)$ as well) is at least $\Omega(n^{1/q}/\varepsilon^2)$. In this subsection, we shall show that $\lambda_{\mathsf{max}}(M_X)$ and $\lambda_{\mathsf{max}}(M_Y)$ can be made at most $O(n^{1/q}/\varepsilon^2)$ as well. While this additional property is not needed for proving Theorem F.5, it shall become useful for proving the desired running time in the next section (see Appendix G).

The following lemma ensures that if we stop the algorithm "whenever we are done", and thus choose possibly less than $n/\varepsilon^2$ matrices, then, $\lambda_{\max}(M_X)$ and $\lambda_{\max}(M_Y)$ can be properly upper bounded.

**Lemma F.6.** *If one stops the algorithm either when $T = \frac{n}{\varepsilon^2}$ iterations are performed, or when the first time that $\sum_{k=0}^{T-1} \mathsf{Dot}(\widehat{L}_{e_k}, X_k)^{1-1/q} \geq \frac{n^{1/q}}{\varepsilon^2}$ is satisfied, then the same result of Theorem F.5 can be obtained, while we have an extra guarantee*

$$\lambda_{\max}(M_X), \lambda_{\max}(M_Y) \leq O\big(\frac{n^{1/q}}{\varepsilon^2}\big) \ .$$

*Proof.* Recall that in the proof of Theorem F.5, we have only used the choice of $T = \frac{n}{\varepsilon^2}$ to deduce (F.5). For this reason, if instead of choosing exactly $T = \frac{n}{\varepsilon^2}$ matrices, we

$$\textit{stop the algorithm at the first time } T \textit{ such that } \sum_{k=0}^{T-1} \mathsf{Dot}(\widehat{L}_{e_k}, X_k)^{1-1/q} \geq \frac{n^{1/q}}{\varepsilon^2} \textit{ is satisfied,}$$

then we automatically have

$$\sum_{k=0}^{T-1} (\widehat{L}_{e_k} \bullet X_k)^{1-1/q} \geq \frac{n^{1/q}}{\varepsilon^2}(1 - O(\varepsilon_1)) \ .$$

Replacing (F.5) with the above lower bound, all results claimed in Theorem F.5 remain true.

In the rest of the proof, we will show that this early termination rule ensures a good upper bound on $\lambda_{\max}(M_X)$ and $\lambda_{\max}(M_Y)$. Indeed, at the time the algorithm is terminated, we must have

$$\sum_{k=0}^{T-1} (\widehat{L}_{e_k} \bullet X_k)^{1-1/q} \leq \sum_{k=0}^{T-1} \mathsf{Dot}(\widehat{L}_{e_k}, X_k)^{1-1/q} \leq \frac{n^{1/q}}{\varepsilon^2} + O(1) \ . \tag{F.9}$$

This is because, since $\widehat{L}_{e_k} \bullet X_k \leq I \bullet X_k = 1$ and thus $\mathsf{Dot}(\widehat{L}_{e_k}, X_k)^{1-1/q} \leq O(1)$, the value $\sum_{k=0}^{T-1} \mathsf{Dot}(\widehat{L}_{e_k}, X_k)^{1-1/q}$ is incremented by at most $O(1)$ at each iteration. As a consequence, at the first iteration it exceeds $n^{1/q}/\varepsilon^2$, the summation must be at least $n^{1/q}/\varepsilon^2 + O(1)$.

Next, substituting (F.9) into (F.3), and choosing $U_X$ so that $M_X \bullet U_X = \lambda_{\max}(M_X)$, we have

$$\lambda_{\max}(M_X) \leq \frac{qn^{1/q}}{(q-1)\alpha} + (1+\xi)\frac{n^{1/q}}{\varepsilon^2} + O(1) = O\big(\frac{n^{1/q}}{\varepsilon^2}\big) \ .$$

Finally, recalling that we have chosen $\mathsf{Dot}(\widehat{L}_{e_k}, X_k) \leq (1+\varepsilon_1)^2 \mathsf{Dot}(\widehat{L}_{e_k}, Y_k)$, this ensures that $(1+\varepsilon_1)^2 M_X \succeq M_Y$. In sum, we obtain that $\lambda_{\max}(M_Y) \leq O(\lambda_{\max}(M_X)) \leq O\big(\frac{n^{1/q}}{\varepsilon^2}\big)$. □

# G Efficient Implementation for Graph Sparsifications

Recall from Appendix F that in order to implement the algorithm described in Theorem F.5, we need to

(C1) Ensure that each $\mathrm{Tr}\widehat{L}_e$ is in $[1-\varepsilon_1, 1]$.

(C2) Compute at each iteration two reals $c^X, c^Y \in \mathbb{R}$ satisfying that $\mathrm{Tr}X_k \in [1, 1+\varepsilon_1]$ and $\mathrm{Tr}Y_k \in [1, 1+\varepsilon_1]$, where

$$X_k \stackrel{\text{def}}{=} \Big(c^X \cdot I - \sum_{j=0}^{k-1} \frac{\alpha \widehat{L}_{e_j}}{\mathsf{Dot}(\widehat{L}_{e_j}, X_j)^{1/q}}\Big)^{-q} \quad \text{and} \quad Y_k \stackrel{\text{def}}{=} \Big(\sum_{j=0}^{k-1} \frac{\alpha \widehat{L}_{e_j}}{\mathsf{Dot}(\widehat{L}_{e_j}, Y_j)^{1/q}} - c^Y \cdot I\Big)^{-q} \ .$$

(C3) Compute at each iteration $\mathsf{Dot}(\widehat{L}_e, X_k)$ and $\mathsf{Dot}(\widehat{L}_e, Y_k)$ which satisfy

$$\widehat{L}_e \bullet X_k \le \mathsf{Dot}(\widehat{L}_e, X_k) \le (1+\varepsilon_1)\widehat{L}_e \bullet X_k \quad \text{and} \quad \widehat{L}_e \bullet Y_k \le \mathsf{Dot}(\widehat{L}_e, Y_k) \le (1+\varepsilon_1)\widehat{L}_e \bullet Y_k \ .$$

In this section, we suppose that we are dealing with a spectral graph sparsification instance (see Appendix B). In other words, we use $I$ to denote $I_{\mathsf{im}(L_G)}$, and have $\widehat{L}_e = \frac{L_G^{-1/2} L_e L_G^{-1/2}}{w_e}$, where $w_e = L_G^{-1} \bullet L_e$ is the effective resistance of edge $e \in [m]$.

Knowing this scaling factor $w_e$ is somewhat important, because we need to ensure that $\mathrm{Tr}\widehat{L}_e$ is between $1-\varepsilon_1$ and 1 according to (C1). Fortunately, Spielman and Srivastava [SS11] have given an algorithm that runs in nearly-linear time, and produces the effective resistances $L_G^{-1} \bullet L_e$ up to a multiplicative error of $1+\varepsilon_1$ for all edges $e \in [m]$, with probability at least $1 - n^{-\Omega(1)}$.

In other words, we can denote by $\widehat{L}_e = \frac{L_G^{-1/2} L_e L_G^{-1/2}}{w_e}$, where each $w_e$ only needs to be between $(1-\varepsilon_1) \cdot L_G^{-1} \bullet L_e$ and $L_G^{-1} \bullet L_e$.

We next wish to show how to implement (C2) and (C3) efficiently. Before that, let us claim that

**Lemma G.1.** *Regardingless of how (C2) and (C3) are implemented, for all iterations,* $c^X, c^Y \le O(\alpha \frac{n^{1/q}}{\varepsilon^2}) = O(\frac{n^{1/q}}{\sqrt{q}\varepsilon})$.

*Proof.* It is first easy to see that $c^Y \le \alpha \cdot \lambda_{\mathsf{max}}(M_Y) \le O(\alpha \frac{n^{1/q}}{\varepsilon^2})$ owing to Lemma F.6. Next, since $\mathrm{Tr}X_k \ge 1$, we must have

$$c^X \le \lambda_{\mathsf{max}}\Big(\sum_{j=0}^{k-1} \frac{\alpha \widehat{L}_{e_j}}{\mathsf{Dot}(\widehat{L}_{e_j}, X_j)^{1/q}}\Big) + n^{1/q} \le \alpha \cdot \lambda_{\mathsf{max}}(M_X) + n^{1/q} \le O(\alpha \frac{n^{1/q}}{\varepsilon^2}) \ . \qquad \square$$

Now, we are ready to prove the main theorem of this section.

**Theorem G.2.** *In an amortized[a] running time of* $\widetilde{O}(\sqrt{q} n^{1/q} m / \varepsilon_1^2 \varepsilon)$ *per iteration, we can implement (C2) and (C3) with probability at least* $1 - n^{-\Omega(1)}$.

*Combining this with the fact that there are at mots* $\frac{n}{\varepsilon^2}$ *iterations, the total running time of our graph sparsification algorithm is*

$$\widetilde{O}\Big(\frac{\sqrt{q} n^{1+1/q} m}{\varepsilon_1^2 \varepsilon^3}\Big) \ .$$

[a] This amortization can be removed, but will result in a slightly more involved implementation to analyze.

Our proof below will make frequent uses of Lemma G.3 and Lemma G.4, two independent lemmas regarding how to efficiently compute matrix inversions of the form $(cI - A)^{-q}$ as well as $(A - cI)^{-q}$. The statements and proofs of these two lemmas are deferred to Appendix G.1.

*Proof.* Both (C2) and (C3) are trivially implementable when $k = 0$, because $X_0 = Y_0 = \frac{1}{n} I$.

Suppose that both of them are implementable at iteration $k-1$. We proceed in 4 steps to prove that they are implementable at iteration $k$ as well.

- **Step I**: prove (C3) for computing $\mathsf{Dot}(\widehat{L}_e, X_k)$.

  Suppose $X_k$ is given in the form of $X_k \stackrel{\text{def}}{=} \Big(c^X \cdot I - \sum_{j=0}^{k-1} \frac{\alpha \widehat{L}_{e_j}}{\mathsf{Dot}(\widehat{L}_{e_j}, X_j)^{1/q}}\Big)^{-q}$ for some $c^X > 0$, and it satisfies $\mathrm{Tr}X_k \in [1, 1+\varepsilon_1]$. (This is done by the inductive assumption.)

Since $\mathrm{Tr} X_k \leq 1+\varepsilon_1 \leq 3/2$, we must have

$$X_k^{-1/q} = c^X \cdot I - \sum_{j=0}^{k-1} \frac{\alpha \widehat{L}_{e_j}}{\mathsf{Dot}(\widehat{L}_{e_j}, X_j)^{1/q}} \succeq \frac{2}{3} I \enspace .$$

This inequality ensures that we can compute $X_k \bullet \widehat{L}_e$ approximately (up to $1+\varepsilon_1$ error) using Lemma G.3. Since $c^X$ is no more than $O(n^{1/q}/\sqrt{q}\varepsilon)$ owing to Lemma G.1, the running time for computing $X_k \bullet \widehat{L}_e$ for all edges $e \in E$ is $\widetilde{O}(c^X qm/\varepsilon_1^2) = \widetilde{O}(\sqrt{q} n^{1/q} m/\varepsilon_1^2 \varepsilon)$.

- **Step II**: prove (C3) for computing $\mathsf{Dot}(\widehat{L}_e, Y_k)$.

  Suppose $Y_k$ is given in the form of $Y_k \stackrel{\text{def}}{=} \Big(\sum_{j=0}^{k-1} \frac{\alpha \widehat{L}_{e_j}}{\mathsf{Dot}(\widehat{L}_{e_j}, Y_j)^{1/q}} - c^Y \cdot I\Big)^{-q}$ for some real $c^Y$, and it satisfies $\mathrm{Tr} Y_k \in [1, 1+\varepsilon_1]$. (This is done by the inductive assumption.) Since $\mathrm{Tr} Y_k \leq 1+\varepsilon_1 \leq 3/2$, we must have

$$Y_k^{-1/q} = \sum_{j=0}^{k-1} \frac{\alpha \widehat{L}_{e_j}}{\mathsf{Dot}(\widehat{L}_{e_j}, Y_j)^{1/q}} - c^Y \cdot I \succeq \frac{2}{3} I \enspace .$$

  This inequality ensures that we can compute $Y_k \bullet \widehat{L}_e$ approximately (up to $1+\varepsilon_1$ error) using Lemma G.4. Since $c^Y$ is no more than $O(n^{1/q}/\sqrt{q}\varepsilon)$ owing to Lemma G.1, the running time for computing $Y_k \bullet \widehat{L}_e$ for all edges $e \in E$ is $\widetilde{O}(c^Y qm/\varepsilon_1^2) = \widetilde{O}(\sqrt{q} n^{1/q} m/\varepsilon_1^2 \varepsilon)$.

- **Step III**: prove (C2) for $X_k$.

  Suppose that $X_{k-1} \stackrel{\text{def}}{=} \big(b^X \cdot I - \sum_{j=0}^{k-2} \frac{\alpha \widehat{L}_{e_j}}{\mathsf{Dot}(\widehat{L}_{e_j}, X_j)^{1/q}}\big)^{-q}$. Since $\mathrm{Tr} X_{k-1} \leq 1+\varepsilon_1 \leq 3/2$, we must have

$$X_{k-1}^{-1/q} = b^X \cdot I - \sum_{j=0}^{k-2} \frac{\alpha \widehat{L}_{e_j}}{\mathsf{Dot}(\widehat{L}_{e_j}, X_j)^{1/q}} \succeq \frac{2}{3} I \enspace .$$

  Recall that we have proved that $X_{k-1}^{-1/q} \succeq \frac{\widehat{L}_{e_j}}{\mathsf{Dot}(\widehat{L}_{e_j}, X_j)^{1/q}}$ (see Claim F.1), combining it with the inequality above and the fact that $\alpha < 1/4$, we have

$$b^X \cdot I - \sum_{j=0}^{k-1} \frac{\alpha \widehat{L}_{e_j}}{\mathsf{Dot}(\widehat{L}_{e_j}, X_j)^{1/q}} \succeq \frac{1}{2} I \enspace . \tag{G.1}$$

  Now, we are ready to perform a binary search to find $c^X$. If one selects $c^X = b^X$, he will get $\mathrm{Tr} X_k \geq \mathrm{Tr} X_{k-1} \geq 1$, and therefore $c^X = b^X$ is a good lower bound for the choice of $c^X$. On the other hand, if one selects $c^X = b^X + n^{1/q}$, he will get $\mathrm{Tr} X_k \leq \mathrm{Tr}(n^{1/q} I)^{-q} = 1$, so $b^X + n^{1/q}$ is a good upper bound for the choice of $c^X$.

  In sum, we can binary search $c^X$ in the interval of $[b^X, b^X + n^{1/q}]$. For each such value of $c^X$ in the process of the binary search, since $c^X$ is no more than $O(n^{1/q}/\sqrt{q}\varepsilon)$ as per Lemma G.1, one can apply Lemma G.3 and approximately compute $\mathrm{Tr}(X_k) = \sum_e X_k \bullet \widehat{L}_e$ up to a multiplicative error of $1+\varepsilon_1$, in time $\widetilde{O}(c^X qm/\varepsilon_1^2) = \widetilde{O}(\sqrt{q} n^{1/q} m/\varepsilon_1^2 \varepsilon)$.

  Since the overhead for the binary search is $\widetilde{O}(1)$, the total running time to compute $c^X$ at an iteration is $\widetilde{O}(\sqrt{q} n^{1/q} m/\varepsilon_1^2 \varepsilon)$.

- **Step IV**: prove (C2) for $Y_k$.

Suppose that $Y_{k-1} \stackrel{\text{def}}{=} \big(\sum_{j=0}^{k-2} \frac{\alpha \widehat{L}_{e_j}}{\mathsf{Dot}(\widehat{L}_{e_j}, Y_j)^{1/q}} - b^Y \cdot I\big)^{-q}$. Since $\mathrm{Tr} Y_{k-1} \leq 1+\varepsilon_1 \leq 3/2$, we must have

$$Y_{k-1}^{-1/q} = \sum_{j=0}^{k-2} \frac{\alpha \widehat{L}_{e_j}}{\mathsf{Dot}(\widehat{L}_{e_j}, Y_j)^{1/q}} - b^Y \cdot I \succeq \frac{2}{3} I \ . \tag{G.2}$$

It is clear from now that it suffices for us to search for $c^Y \geq b^Y$, because if one selects $c^Y = b^Y$, he will get $\mathrm{Tr} Y_k \leq \mathrm{Tr} Y_{k-1} \leq 1+\varepsilon_1$, and therefore $c^Y = b^Y$ is a good lower bound. However, unlike Step III, one cannot perform a simple binary search on $c^Y$ because there is no good upper bound for $c^Y$.[12]

Instead, consider the following increment-and-binary-search algorithm. Beginning from $b^Y$, we first choose $c^Y = b^Y + \frac{1}{6}$. This choice of $c^Y$ ensures that, according to (G.2),

$$Y_k^{-1/q} = \sum_{j=0}^{k-1} \frac{\alpha \widehat{L}_{e_j}}{\mathsf{Dot}(\widehat{L}_{e_j}, Y_j)^{1/q}} - c^Y \cdot I \succeq \frac{1}{2} I \ .$$

Therefore, we can compute $\mathrm{Tr}(Y_k) = \sum_e Y_k \bullet \widehat{L}_e$ approximately using Lemma G.4. If the approximation computation from Lemma G.4 tells us that $\mathrm{Tr}(Y_k) \geq 1$, we stop the increment of $c^Y$. Otherwise, we conclude that $\mathrm{Tr}(Y_k)$ is still less than or equal to $1+\varepsilon_1$, and continue to try $c^Y = b^T + \frac{i}{6}$ for $i = 2, 3, 4, \ldots$. We stop this increment until we find some integer $i$ so that $\mathrm{Tr}(Y_k) \geq 1$.

At this moment, we have that

$$\mathrm{Tr}\Big(\sum_{j=0}^{k-1} \frac{\alpha \widehat{L}_{e_j}}{\mathsf{Dot}(\widehat{L}_{e_j}, Y_j)^{1/q}} - (b^Y + \frac{i-1}{6}) \cdot I\Big)^{-q} \leq 1+\varepsilon_1 \quad \text{and}$$

$$\mathrm{Tr}\Big(\sum_{j=0}^{k-1} \frac{\alpha \widehat{L}_{e_j}}{\mathsf{Dot}(\widehat{L}_{e_j}, Y_j)^{1/q}} - (b^Y + \frac{i}{6}) \cdot I\Big)^{-q} \geq 1 \ .$$

Therefore, we can perform a binary search for $c^Y$ between $b^Y + \frac{i-1}{6}$ and $b^Y + \frac{i}{6}$ for, and in $\widetilde{O}(1)$ time we can find some value in this interval which satisfies $\mathrm{Tr}(Y_k) \in [1, 1+\varepsilon_1]$.

Again, since we always have $c^Y \leq O(n^{1/q}/\sqrt{q}\varepsilon)$ owing to Lemma G.1, the binary search step costs a running time that is at most $\widetilde{O}(c^Y qm/\varepsilon_1^2) = \widetilde{O}(\sqrt{q} n^{1/q} m/\varepsilon_1^2 \varepsilon)$ owing to Lemma G.4.

The incrementation procedure takes a running time $\widetilde{O}(\sqrt{q} n^{1/q} m/\varepsilon_1^2 \varepsilon)$ for each increment of $\frac{1}{6}$. However, throughout the algorithm, we increment $c^Y$ by $1/6$ at most $O(n^{1/q}/\sqrt{q}\varepsilon)$ times in total as per Lemma G.1. This running time, after amortization, is going to be dominated by that of the binary search.

Overall, we have shown that (C2) and (C3) can be implemented to run in $\widetilde{O}(\sqrt{q} n^{1/q} m/\varepsilon_1^2 \varepsilon)$ time (in amortization) per iteration. Since there are a total of at most $\frac{n}{\varepsilon^2}$ iterations, the desired running time is obtained. □

## G.1 Missing Lemmas

In this subsection, we state and prove Lemma G.3 and Lemma G.4 for the efficient computations of the matrix inverses needed for the previous subsection.

[12] In fact, if one is allowed to compute the smallest eigenvalue of $\sum_{j=0}^{k-1} \frac{\alpha \widehat{L}_{e_j}}{\mathsf{Dot}(\widehat{L}_{e_j}, Y_j)^{1/q}}$, he can perform a binary search as described in Section 6. However, we have chosen not to implement that algorithm because the running time analysis for the max/min eigenvalue computation is only longer than the current one.

**Lemma G.3.** *Suppose that we are given positive reals* $c$ *and* $s_0, \dots, s_{k-1}$ *satisfying* $cI - \sum_{j=0}^{k-1} s_j \check{L}_{e_j} \succeq \frac{1}{2}I$*, where each* $\check{L}_e$ *is the normalized edge Laplacian and* $k = O(m)$*. Let* $q$ *be any positive even integer. Then, we can compute a matrix* $T \in \mathbb{R}^{m' \times m}$ *in time* $\widetilde{O}(cqm/\varepsilon_1^2)$*, where* $T$ *has* $m' = \Theta(\log n/\varepsilon_1^2)$ *rows and satisfies that, with probability at least* $1 - n^{-\Omega(1)}$*,*

$$\forall e \in E, \qquad X \bullet \check{L}_e \leq \|T\chi_e\|_2^2 \leq (1+\varepsilon_1) X \bullet \check{L}_e \ , \qquad \text{where } X \stackrel{\text{def}}{=} \Big(cI - \sum_{j=0}^{k-1} s_j \check{L}_{e_j}\Big)^{-q} \quad .$$

**Lemma G.4.** *Suppose we are given positive* $s_0, \dots, s_{k-1}$ *and a possibly negative real* $c$ *satisfying that* $\sum_{j=0}^{k-1} s_j \check{L}_{e_j} - cI \succeq \frac{1}{2}I$*, where each* $\check{L}_e$ *is the normalized edge Laplacian and* $k = O(m)$*. Let* $q$ *be any positive even integer. Then, we can compute a matrix* $T \in \mathbb{R}^{m' \times m}$ *in time* $\widetilde{O}(cqm/\varepsilon_1^2)$*, where* $T$ *has* $m' = \Theta(\log n/\varepsilon_1^2)$ *rows and satisfies that, with probability at least* $1 - n^{-\Omega(1)}$*,*

$$\forall e \in E, \qquad Y \bullet \check{L}_e \leq \|T\chi_e\|_2^2 \leq (1+\varepsilon_1) Y \bullet \check{L}_e \ , \qquad \text{where } Y \stackrel{\text{def}}{=} \Big(\sum_{j=0}^{k-1} s_j \check{L}_{e_j} - cI\Big)^{-q} \quad .$$

Our proofs to the above lemmas rely on the following auxiliary tools.

### G.1.1 Auxiliary Tools

The first one is the famous Laplacian linear system solver, written in the matrix language.

**Theorem G.5.** *For parameter* $\alpha \in [0,1]$*. Given any Laplacian matrix* $L$ *that corresponds to a graph with* $m$ *edges, there exist an approximation* $\overline{L}^{-1}$ *which satisfies that, with probability at least* $1 - n^{-\Omega(1)}$*,* $(1-\delta)L^{-1} \preceq \overline{L}^{-1} \preceq (1+\delta)L^{-1}$*, and for every vector* $v \in \mathbb{R}^n$*,* $\overline{L}^{-1}v$ *can be computed in time* $\widetilde{O}(m \log(1/\delta))$*.*

*Proof.* The algorithms presented in [ST04] can be expressed as matrices $\overline{L}^{-1}$ which satisfy that, with high probability, for every $x \in \mathbb{R}^n$, the vectors $L^{-1}x$ and $\overline{L}^{-1}$ are close under the so-called $L$-norm, or in symbols, $\|\overline{L}^{-1}x - L^{-1}x\|_L^2 \leq \delta^2 \|L^{-1}x\|_L^2$. After expanding this out using the definition of the $L$-norm, we have

$$\begin{aligned} & x^T\big(\overline{L}^{-1} - L^{-1}\big)L\big(\overline{L}^{-1} - L^{-1}\big)x \leq \delta^2 \cdot x^T L^{-1} L L^{-1} x \\ \Longrightarrow \quad & \big(\overline{L}^{-1} - L^{-1}\big)L\big(\overline{L}^{-1} - L^{-1}\big) \preceq \delta^2 \cdot L^{-1} \\ \Longrightarrow \quad & (L^{1/2}\overline{L}^{-1}L^{1/2} - I)^2 \preceq \delta^2 I \\ \Longrightarrow \quad & -\delta I \preceq L^{1/2}\overline{L}^{-1}L^{1/2} - I \preceq \delta I \\ \Longrightarrow \quad & (1-\delta)L^{-1} \preceq \overline{L}^{-1} \preceq (1+\delta)L^{-1} \ . \end{aligned}$$

The running time $\widetilde{O}(m \log(1/\delta))$ follows from that of [ST04]. □

The next two lemmas are the classical results on approximating $(I-A)^{-q}$ and $(A-I)^{-q}$ using Taylor expansions.

**Lemma G.6.** *The polynomial* $\mathsf{P}(A) = I + A + \cdots + A^{d-1}$ *satisfies that for all* $0 \preceq A \preceq (1-\delta)I$*,*

$$0 \preceq (I-A)^{-1} - \mathsf{P}(A) \preceq (1-\delta)^d \cdot (I-A)^{-1} \ .$$

*As a consequence, for every integer* $q \geq 1$*,*

$$(1 - q(1-\delta)^d) \cdot (I-A)^{-q} \preceq \mathsf{P}^q(A) \preceq (I-A)^{-q} \ .$$

*Proof.* We first note that for every $x \in [0, 1-\delta]$, we have

$$0 \le \frac{1}{1-x} - (1 + x + \cdots + x^{d-1}) = x^d + x^{d+1} + \cdots = \frac{x^d}{1-x} \le \frac{(1-\delta)^d}{1-x} \quad . \tag{G.3}$$

As a consequence, we have that

$$0 \preceq (I-A)^{-1} - (1 + A + \cdots + A^{d-1}) \preceq (1-\delta)^d \cdot (I-A)^{-1} \enspace ,$$

which can be proved by first assuming (without loss of generality) that $A$ is diagonal, and then analyzing each diagonal entry using (G.3).

To prove the result for $(I-A)^{-q}$, we first notice that $(I-A)^{-1}$ and $\mathsf{P}(A)$ are commutable. Therefore, $\mathsf{P}(A) \preceq (I-A)^{-1}$ directly implies $\mathsf{P}^q(A) \preceq (I-A)^{-q}$, which gives one side of the inequality. To see the other side, we rewrite

$$(1 - (1-\delta)^d) \cdot (I-A)^{-1} \preceq \mathsf{P}(A) \enspace ,$$

and then take the $q$-th power on both sides. This yields

$$\big(1 - q(1-\delta)^d\big) \cdot (I-A)^{-q} \preceq (1 - (1-\delta)^d)^q \cdot (I-A)^{-q} \preceq \mathsf{P}^q(A) \enspace ,$$

which finishes the proof of the lemma. □

**Lemma G.7.** *The polynomial* $\mathsf{P}(A) = A + A^2 + \cdots + A^d$ *satisfies that for all* $(1+\delta)I \preceq A$*,*

$$0 \preceq (A-I)^{-1} - \mathsf{P}(A^{-1}) \preceq (1+\delta)^{-d} \cdot (A-I)^{-1} \enspace .$$

*As a consequence, for every integer* $q \ge 1$*,*

$$(1 - q(1+\delta)^{-d}) \cdot (A-I)^{-q} \preceq \mathsf{P}^q(A^{-1}) \preceq (A-I)^{-q} \enspace .$$

*Proof.* We first note that for every $x \ge 1+\delta$, we have

$$0 \le \frac{1}{x-1} - (x^{-1} + x^{-2} + \cdots + x^{-d}) = x^{-d-1} + x^{-d-2} + \cdots = \frac{1}{x^d}\frac{1}{x-1} \le \frac{1}{(1+\delta)^d}\frac{1}{x-1} \quad . \tag{G.4}$$

As a consequence, we have that

$$0 \preceq (A-I)^{-1} - (A^{-1} + A^{-2} + \cdots + A^{-d}) \preceq (1+\delta)^{-d} \cdot (A-I)^{-1} \enspace ,$$

which can be proved by first assuming (without loss of generality) that $A$ is diagonal, and then analyzing each diagonal entry using (G.4).

To prove the result for $(A-I)^{-q}$, we first notice that $(A-I)^{-1}$ and $\mathsf{P}(A^{-1})$ are commutable. Therefore, $\mathsf{P}(A^{-1}) \preceq (A-I)^{-1}$ directly implies $\mathsf{P}^q(A^{-1}) \preceq (A-I)^{-q}$, which gives one side of the inequality. To see the other side, we rewrite

$$(1 - (1+\delta)^{-d}) \cdot (A-I)^{-1} \preceq \mathsf{P}(A^{-1}) \enspace ,$$

and then take the $q$-th power on both sides. This yields

$$\big(1 - q(1+\delta)^{-d}\big) \cdot (A-I)^{-q} \preceq (1 - (1+\delta)^{-d})^q \cdot (A-I)^{-q} \preceq \mathsf{P}^q(A^{-1}) \enspace ,$$

which finishes the proof of the lemma. □

### G.1.2 Missing Proofs of Lemma G.3 and G.4

**Lemma G.3.** *Suppose that we are given positive reals* $c$ *and* $s_0, \dots, s_{k-1}$ *satisfying* $cI - \sum_{j=0}^{k-1} s_j \check{L}_{e_j} \succeq \frac{1}{2}I$*, where each* $\check{L}_e$ *is the normalized edge Laplacian and* $k = O(m)$*. Let* $q$ *be any positive even integer. Then, we can compute a matrix* $T \in \mathbb{R}^{m' \times m}$ *in time* $\widetilde{O}(cqm/\varepsilon_1^2)$*, where* $T$ *has* $m' = \Theta(\log n/\varepsilon_1^2)$ *rows and satisfies that, with probability at least* $1 - n^{-\Omega(1)}$*,*

$$\forall e \in E, \qquad X \bullet \check{L}_e \le \|T\chi_e\|_2^2 \le (1+\varepsilon_1) X \bullet \check{L}_e \enspace , \qquad \textit{where } X \stackrel{\text{def}}{=} \Big(cI - \sum_{j=0}^{k-1} s_j \check{L}_{e_j}\Big)^{-q} \quad .$$

*Proof.* Denoting by $A = \frac{1}{c}\sum_{j=0}^{k-1} s_j \check{L}_{e_j}$, we have $0 \preceq A \preceq (1-\frac{1}{2c})I$ by the assumption. Now we apply Lemma G.6, and let $\mathsf{P}(A)$ be the matrix polynomial of degree $d = \Theta(c\log(q/\varepsilon_1))$ from Lemma G.6. By the approximation guarantee, we have for every edge $e \in E$,

$$X \bullet \check{L}_e = \Big(cI - \sum_{j=0}^{k-1} s_j \check{L}_{e_j}\Big)^{-q} \bullet \check{L}_e = \Big(1 \pm \frac{\varepsilon_1}{10}\Big) \cdot c^{-q} \cdot \mathsf{P}^q(A) \bullet \check{L}_e \ . \tag{G.5}$$

Therefore, it suffices for us to compute $\mathsf{P}^q(A) \bullet \check{L}_e$ for each possible edge $e$.

Next, let $\overline{L_G}^{-1}$ be the approximation of $L_G^{-1}$ from Theorem G.5 that satisfies

$$(1 - \frac{\varepsilon_1}{10dq})L_G^{-1} \preceq \overline{L_G}^{-1} \preceq (1 + \frac{\varepsilon_1}{10dq})L_G^{-1} \ .$$

Denoting by $L_s \stackrel{\text{def}}{=} \sum_{j=0}^{k-1} \frac{s_j}{c} L_{e_j}$, we have $A = L_G^{-1/2} L_s L_G^{-1/2}$. Accordingly, for every edge $e \in E$,

$$\begin{aligned}
\mathsf{P}^q(A) \bullet \check{L}_e &= \mathrm{Tr}\Big(\mathsf{P}^q\big(L_G^{-1/2} L_s L_G^{-1/2}\big)\ L_G^{-1/2} L_e L_G^{-1/2}\Big) \\
&= \mathrm{Tr}\Big(\mathsf{P}^q\big(L_G^{-1} L_s\big)\ L_G^{-1} L_e\Big) \\
&= \mathrm{Tr}\Big(\mathsf{P}^{q/2}\big(L_G^{-1} L_s\big)\ L_G^{-1}\ \mathsf{P}^{q/2}\big(L_s L_G^{-1}\big) L_e\Big) \\
&= \mathrm{Tr}\Big(\mathsf{P}^{q/2}\big(L_G^{-1} L_s\big)\ L_G^{-1}\ B^T W B^T\ L_G^{-1}\ \mathsf{P}^{q/2}\big(L_s L_G^{-1}\big) L_e\Big) \\
&\stackrel{①}{=} (1 \pm \varepsilon_1/10) \cdot \mathrm{Tr}\Big(\mathsf{P}^{q/2}\big(\overline{L_G}^{-1} L_s\big)\ \overline{L_G}^{-1}\ B^T W B^T\ \overline{L_G}^{-1}\ \mathsf{P}^{q/2}\big(L_s \overline{L_G}^{-1}\big) L_e\Big) \\
&= (1 \pm \varepsilon_1/10) \cdot w_e \cdot \chi_e^T \mathsf{P}^{q/2}\big(\overline{L_G}^{-1} L_s\big)\ \overline{L_G}^{-1}\ B^T W B^T\ \overline{L_G}^{-1}\ \mathsf{P}^{q/2}\big(L_s \overline{L_G}^{-1}\big)\chi_e \\
&= (1 \pm \varepsilon_1/10) \cdot w_e \cdot \Big\|W^{1/2} B^T \overline{L_G}^{-1} \mathsf{P}^{q/2}\big(L_s \overline{L_G}^{-1}\big)\chi_e\Big\|_2^2 \ .
\end{aligned} \tag{G.6}$$

Above, ① follows because each $\overline{L_G}^{-1}$ is a $(1 \pm \frac{\varepsilon_1}{10dq})$ approximation to $L_G^{-1}$, while we have at most $(d-1)q + 2 \le dq$ copies of $L_G^{-1}$ in any sequence of the matrix multiplication on the left hand side of ①.

For this reason, we can preprocess by computing $T' \stackrel{\text{def}}{=} Q W^{1/2} B^T \overline{L_G}^{-1} \mathsf{P}^{q/2}\big(L_s \overline{L_G}^{-1}\big) \in \mathbb{R}^{m' \times n}$, where $Q \in \mathbb{R}^{m' \times m}$ is some Johnson-Lindenstrauss random matrix with $m' = \Theta(\log n/\varepsilon_1^2)$ rows. This matrix $T'$ satisfies that, with probability at least $1 - O(n^{-\Omega(1)})$,

$$\forall e \in E, \qquad \Big\|Q W^{1/2} B^T \overline{L_G}^{-1} \mathsf{P}^{q/2}\big(L_s \overline{L_G}^{-1}\big)\chi_e\Big\|_2^2 = (1 \pm \varepsilon_1/10)\|T'\chi_e\|_2^2 \ . \tag{G.7}$$

Combining (G.5), (G.6), and (G.7) together, we have

$$\forall e \in E, \qquad X \bullet \check{L}_e = (1 \pm \varepsilon_1/3) \cdot c^{-q} \cdot w_e \cdot \|T'\chi_e\|_2^2 \ .$$

Defining $T \stackrel{\text{def}}{=} \big(\frac{1}{1-\varepsilon_1/3} \cdot c^{-q} \cdot w_e\big)^{1/2} \cdot T'$, we get the desired inequality in Lemma G.3.

Finally, we emphasize that the above computation of $T$ requires $\widetilde{O}(dq \cdot m' \cdot m) = \widetilde{O}(cqm/\varepsilon_1^2)$ time. This is because, each row of $T$ can be computed by left multiplying each row of $Q$ with the matrix $W^{1/2} B^T \overline{L_G}^{-1} \mathsf{P}^{q/2}\big(L_s \overline{L_G}^{-1}\big)$.[13] The running time now follows from (i) we need to compute vector-matrix multiplication $O(dq)$ times, which is the power of the polynomial $\mathsf{P}^{q/2}(\cdot)$, and (ii)

[13]This can be implemented as follows. For any row vector of $Q$, denote it by $u^T \in \mathbb{R}^m$. We first sequentially compute

- $v^T \leftarrow u^T W^{1/2}$,
- $v^T \leftarrow v^T B^T$, and
- $v^T \leftarrow v^T \overline{L_G}^{-1}$.

Now, suppose $\mathsf{P}^{q/2}\big(L_s \overline{L_G}^{-1}\big) = \sum_{i=0}^{dq/2} c_i \big(L_s \overline{L_G}^{-1}\big)^i$ where each $c_i$ is the coefficient of the $i$-th power term. We continue and compute

- $w^T \leftarrow \vec{0}$.

Theorem G.5 implies that for inversion $v^T\overline{L_G}^{-1}$ can be computed in time $\widetilde{O}(m\log(dq/\varepsilon_1))$ for any vector $v$. □

**Lemma G.4.** *Suppose we are given positive $s_0,\dots,s_{k-1}$ and a possibly negative real $c$ satisfying that $\sum_{j=0}^{k-1} s_j\check{L}_{e_j} - cI \succeq \frac{1}{2}I$, where each $\check{L}_e$ is the normalized edge Laplacian and $k=O(m)$. Let $q$ be any positive even integer. Then, we can compute a matrix $T\in\mathbb{R}^{m'\times m}$ in time $\widetilde{O}(cqm/\varepsilon_1^2)$, where $T$ has $m'=\Theta(\log n/\varepsilon_1^2)$ rows and satisfies that, with probability at least $1-n^{-\Omega(1)}$,*

$$\forall e\in E,\qquad Y\bullet\check{L}_e \le \|T\chi_e\|_2^2 \le (1+\varepsilon_1)Y\bullet\check{L}_e\ ,\qquad \text{where } Y \stackrel{\text{def}}{=} \Big(\sum_{j=0}^{k-1} s_j\check{L}_{e_j} - cI\Big)^{-q}\ .$$

*Proof.* There are two cases: $c>0$ or $c\le 0$. We begin with the case when $c>0$.

Denoting by $A=\frac{1}{c}\sum_{j=0}^{k-1}s_j\check{L}_{e_j}$, we have $A\succeq(1+\frac{1}{2c})I$ by the assumption. Now we apply Lemma G.7, and let $\mathsf{P}(A)$ be the matrix polynomial of degree $d=\Theta(c\log(q/\varepsilon_1))$ from Lemma G.7. By the approximation guarantee, we have for every edge $e\in E$,

$$Y\bullet\check{L}_e = \Big(\sum_{j=0}^{k-1} s_j\check{L}_{e_j} - cI\Big)^{-q}\bullet\check{L}_e = \Big(1\pm\frac{\varepsilon_1}{10}\Big)\cdot c^{-q}\cdot\mathsf{P}^q(A^{-1})\bullet\check{L}_e\ . \tag{G.8}$$

Therefore, it suffices for us to compute $\mathsf{P}^q(A^{-1})\bullet\check{L}_e$ for each possible edge $e$.

Denoting by $L_s\stackrel{\text{def}}{=}\sum_{j=0}^{k-1}\frac{s_j}{c}L_{e_j}$, we have $A^{-1}=L_G^{1/2}L_s^{-1}L_G^{1/2}$. Next, let $\overline{L_s}^{-1}$ and $\overline{L_G}^{-1}$ respectively be the approximation of $L_s^{-1}$ and $L_G^{-1}$ from Theorem G.5 that satisfy

$$(1-\frac{\varepsilon_1}{10dq})L_s^{-1}\preceq\overline{L_s}^{-1}\preceq(1+\frac{\varepsilon_1}{10dq})L_s^{-1}\ ,\ \text{and}$$
$$(1-\frac{\varepsilon_1}{10dq})L_G^{-1}\preceq\overline{L_G}^{-1}\preceq(1+\frac{\varepsilon_1}{10dq})L_G^{-1}\ .$$

Accordingly, for every edge $e\in E$,

$$\begin{aligned}
\mathsf{P}^q(A^{-1})\bullet\check{L}_e &= \mathrm{Tr}\Big(\mathsf{P}^q\big(L_G^{1/2}L_s^{-1}L_G^{1/2}\big)\ L_G^{-1/2}L_eL_G^{-1/2}\Big)\\
&= \mathrm{Tr}\Big(\mathsf{P}^q\big(L_s^{-1}L_G\big)\ L_G^{-1}L_e\Big)\\
&= \mathrm{Tr}\Big(\mathsf{P}^{q/2}\big(L_s^{-1}L_G\big)\ L_G^{-1}\ \mathsf{P}^{q/2}\big(L_GL_s^{-1}\big)L_e\Big)\\
&= \mathrm{Tr}\Big(\mathsf{P}^{q/2}\big(L_s^{-1}L_G\big)\ L_G^{-1}\ B^TWB^T\ L_G^{-1}\ \mathsf{P}^{q/2}\big(L_GL_s^{-1}\big)L_e\Big)\\
&\stackrel{①}{=} (1\pm\varepsilon_1/10)\cdot\mathrm{Tr}\Big(\mathsf{P}^{q/2}\big(\overline{L_s}^{-1}L_G\big)\ \overline{L_G}^{-1}\ B^TWB^T\ \overline{L_G}^{-1}\ \mathsf{P}^{q/2}\big(L_G\overline{L_s}^{-1}\big)L_e\Big)\\
&= (1\pm\varepsilon_1/10)\cdot w_e\cdot\chi_e^T\mathsf{P}^{q/2}\big(\overline{L_s}^{-1}L_G\big)\ \overline{L_G}^{-1}\ B^TWB^T\ \overline{L_G}^{-1}\ \mathsf{P}^{q/2}\big(L_G\overline{L_s}^{-1}\big)\chi_e\\
&= (1\pm\varepsilon_1/10)\cdot w_e\cdot\Big\|W^{1/2}B^T\overline{L_G}^{-1}\mathsf{P}^{q/2}\big(L_G\overline{L_s}^{-1}\big)\chi_e\Big\|_2^2
\end{aligned}\tag{G.9}$$

Above, ① follows because each $\overline{L_s}^{-1}$ (resp. $\overline{L_G}^{-1}$) is a $(1\pm\frac{\varepsilon_1}{10dq})$ approximation to $L_s^{-1}$ (resp. $L_G^{-1}$), while we have at most $(d-1)q+2\le dq$ copies of $L_s^{-1}$ and $L_G^{-1}$ in any sequence of the matrix multiplication on the left hand side of ①.

---

- For $i\leftarrow 0$ to $dq/2$,
  - $w^T\leftarrow w^T+v^T$.
  - $v^T\leftarrow v^TL_s$.
  - $v^T\leftarrow v^T\overline{L_G}^{-1}$.

In the end, the value of the row vector $w^T$ is precisely the desired $u^TW^{1/2}B^T\overline{L_G}^{-1}\mathsf{P}^{q/2}(L_s\overline{L_G}^{-1})$.

For this reason, we can preprocess by computing $T' \stackrel{\text{def}}{=} QW^{1/2}B^T\overline{L_G}^{-1}\mathsf{P}^{q/2}\big(L_G\overline{L_s}^{-1}\big) \in \mathbb{R}^{m'\times n}$, where $Q \in \mathbb{R}^{m'\times m}$ is some Johnson-Lindenstrauss random matrix with $m' = \Theta(\log n/\varepsilon_1^2)$ rows. This matrix $T'$ satisfies that, with probability at least $1 - O(n^{-\Omega(1)})$,

$$\forall e \in E, \qquad \Big\|QW^{1/2}B^T\overline{L_G}^{-1}\mathsf{P}^{q/2}\big(L_G\overline{L_s}^{-1}\big)\chi_e\Big\|_2^2 = (1 \pm \varepsilon_1/10)\|T'\chi_e\|_2^2 \ . \tag{G.10}$$

Combining (G.8), (G.9), and (G.10), we have

$$\forall e \in E, \qquad Y \bullet \check{L}_e = (1 \pm \varepsilon_1/3) \cdot c^{-q} \cdot w_e \cdot \|T'\chi_e\|_2^2 \ .$$

Defining $T \stackrel{\text{def}}{=} \big(\frac{1}{1-\varepsilon_1/3} \cdot c^{-q} \cdot w_e\big)^{1/2} \cdot T'$, we get the desired inequality in Lemma G.4.

Finally, we emphasize that the computation of $T$ requires $\widetilde{O}(dq \cdot m' \cdot m) = \widetilde{O}(dqm/\varepsilon_1^2)$ time. This is because, each row of $T$ can be computed by left multiplying each row of $Q$ with the matrix $W^{1/2}B^T\overline{L_G}^{-1}\mathsf{P}^{q/2}\big(L_G\overline{L_s}^{-1}\big)$.[14] The running time now follows from (i) we need to compute vector-matrix multiplication $O(dq)$ times, which is the power of the polynomial $\mathsf{P}^{q/2}(\cdot)$, and (ii) Theorem G.5 implies the inversions $v^T\overline{L_G}^{-1}$ and $v^T\overline{L_s}^{-1}$ can both be computed in time $\widetilde{O}(m\log(dq/\varepsilon_1))$, for any vector $v$.

In the second case, if $c \le 0$, we can write

$$Y = \Big(\sum_{j=0}^{k-1} s_j\check{L}_{e_j} - cI\Big)^{-q} = \Big(L_G^{-1/2}(L_s - cL_G)L_G^{-1/2}\Big)^{-q} \ .$$

Therefore, denoting by $L'_s = L_s - cL_G$, which is another graph Laplacian matrix (with positive edge weights), we can write

$$\begin{aligned}
Y \bullet \check{L}_e &= \mathrm{Tr}\Big(\big(L_G^{-1/2}L'_sL_G^{-1/2}\big)^{-q}\ L_G^{-1/2}L_eL_G^{-1/2}\Big) \\
&= \mathrm{Tr}\Big(\big(L_s'^{-1}L_G\big)^{-q/2}\ L_G^{-1}\ \big(L_GL_s'^{-1}\big)^{-q/2}L_e\Big) \\
&= w_e \cdot \chi_e^T\big(L_s'^{-1}L_G\big)^{-q/2}\ L_G^{-1}B^TWBL_G^{-1}\ \big(L_GL_s'^{-1}\big)^{-q/2}\chi_e \\
&= w_e \cdot \big\|W^{1/2}BL_G^{-1}\ \big(L_GL_s'^{-1}\big)^{-q/2}\chi_e\big\|_2^2 \ .
\end{aligned}$$

It is now clear that similar to the previous case, we can approximately compute $L_s'^{-1}$ and $L_G^{-1}$ using Theorem G.5, and apply the Johnson-Lindenstrauss dimension reduction. We skip the detailed proofs here because it is only a repetition. □

# H Efficient Implementation for Other Problems

As we have seen in Appendix G, Lemma G.3 and Lemma G.4 are at the core of our efficient implementation for the graph sparsification problem. For each other possible sparsification problem, as long as these two lemmas can be properly revised, we can also obtain fast running times. Let us illustrate how to obtain such running times for two applications below.

**Sparsifying sums of rank-1 matrices.** To solve the problem in Theorem 2, it is not hard to verify that Lemma G.3 can be revised as follows:

*Suppose that we are given positive reals $c$ and $s_0, \dots, s_{k-1}$ satisfying $cI - \sum_{j=0}^{k-1} s_j\widehat{L}_{e_j} \succeq \frac{1}{2}I$, where each $\widehat{L}_{e_j} = v_{e_j}v_{e_j}^T$ is an explicit $n\times n$ rank-1 matrix and $k = O(m)$. Let $q$ be any positive even integer. Then, we can compute a matrix $T \in \mathbb{R}^{m'\times n}$ in time $\widetilde{O}(cqn^2/\varepsilon_1^2)$, where $T$ has $m' =$*

[14] This can be implemented in a similar manner as discussed in Footnote 13.

$\Theta(\log n/\varepsilon_1^2)$ *rows and satisfies that, with probability at least* $1-n^{-\Omega(1)}$,

$$\forall e\in E, \qquad X\bullet \widehat{L}_e \leq \|Tv_e\|_2^2 \leq (1+\varepsilon_1)X\bullet\widehat{L}_e \enspace, \qquad \textit{where } X \stackrel{\text{def}}{=} \Big(cI-\sum_{j=0}^{k-1}s_j\widehat{L}_{e_j}\Big)^{-q} \enspace .$$

The key idea for proving the above variant of Lemma G.3 is to note that the matrix inequality $cI-\sum_{j=0}^{k-1}s_j\widehat{L}_{e_j}\succeq\frac{1}{2}I$ implies that the condition number for PSD matrix $M\stackrel{\text{def}}{=}cI-\sum_{j=0}^{k-1}s_j\widehat{L}_{e_j}$ is at most $O(c)$. Therefore, one can use for instance steepest descent (or even conjugate gradient or Chebyshev method) to compute $M^{-1}v$ in time $O(cn^2)$ for every vector $v\in\mathbb{R}^n$. Next, one can apply the similar Johnson-Lindenstrauss dimension reduction as presented in the proof of Lemma G.3.

A similar variant of Lemma G.4 can be proved similarly.

In sum, each iteration of our Appendix F is dominated by the computational time need to (1) compute the matrix $T\in\mathbb{R}^{m'\times n}$, which takes time $\widetilde{O}(cqn^2/\varepsilon_1^2)=\widetilde{O}(\sqrt{q}n^{2+1/q}/\varepsilon\varepsilon_1^2)$, and (2) compute $Tv_e$ for all $e\in[m]$, which takes time $O(mn/\varepsilon_1^2)$. Taking into account that we have $T=n/\varepsilon^2$ such iterations, this is a total running time of

$$O\Big(\frac{\sqrt{q}n^{3+1/q}}{\varepsilon^2\varepsilon_1^2}+\frac{mn^2}{\varepsilon_1^2\varepsilon^2}\Big) \enspace .$$

**Subgraph sparsification.** Given a weighted undirected graph $G$ that can be decomposed into edge-disjoint subgraphs, the goal of linear-sized subgraph sparsification is to construct a $(1+O(\varepsilon))$-spectral sparsifier $G'$ to $G$, so that $G'$ consists only of the reweighted versions of at most $n/\varepsilon^2$ given subgraphs.

In symbols, suppose that the edges of some weighted undirected graph $G$ of $n$ vertices and $m'$ edges are decomposed into a disjoint union $E=\biguplus_{i=1}^m E_i$. We are interested in finding scalars $s_e\geq 0$ with $|\{e: s_e>0\}|\leq O(n/\varepsilon^2)$ such that, letting $L\stackrel{\text{def}}{=}\sum_{e=1}^m s_e\cdot L_{G[E_e]}$, where $L_{E_e}$ is the graph Laplacian matrix on the subgraph of $G$ induced by $E_e$, we have $L_G\preceq L\preceq(1+\varepsilon)L_G$.

For this sparsification problem, for each $e\in[m]$, we define $\widehat{L}_e=\frac{L_G^{-1/2}L_{G[E_e]}L_G^{-1/2}}{w_e}$ to be the normalized subgraph Laplacian scaled by $w_e$. Here, $w_e$ is the scaling parameter which ensures that $\mathrm{Tr}\widehat{L}_e$ is between $1-\varepsilon_1$ and 1. (It suffices to compute $L_G^{-1}\bullet L_{G[E_e]}$ up to a multiplicative $1+\varepsilon_1$ error, and then assign $w_e\approx L_G^{-1}\bullet L_{G[E_e]}$.)

For this particular problem, we do not even need to revise Lemma G.3 or Lemma G.4. Recall that we only need to compute 'matrix inversions' of the form

$$\Big(c^X\cdot I-\sum_{j=0}^{k-1}\frac{\alpha\widehat{L}_{e_j}}{\mathsf{Dot}(\widehat{L}_{e_j},X_j)^{1/q}}\Big)^{-q}\bullet\widehat{L}_e \enspace ,$$

while each $\widehat{L}_{e_j}$ is now —instead of a single (scaled) edge Laplacian matrix— the summation of a few (scaled) edge Laplacian matrices. This remains to be the same problem Lemma G.3 is trying to implement. The total running time for this subgraph sparsification is therefore

$$\widetilde{O}\Big(\frac{\sqrt{q}n^{1+1/q}m'}{\varepsilon_1^2\varepsilon^3}\Big) \enspace .$$